\documentclass[runningheads]{llncs}
\usepackage{graphicx}
 
\newcommand\ie{\emph{i.e.}}

\newcommand\etal{\emph{et al.}}
\usepackage{tikz}
\usepackage{comment}
\usepackage{amsmath,amssymb} 
\usepackage{epsfig}
\usepackage[toc,page]{appendix}
\usepackage{graphicx}
\usepackage{amsmath}
\usepackage{amssymb}
\usepackage{color}
\usepackage{multicol, blindtext}
\usepackage{booktabs}
\usepackage{soul}
\usepackage{csquotes}
\usepackage{multirow}
\usepackage{cancel}
\usepackage[accsupp]{axessibility}  
\newcommand{\Rows}[1]{\multirow{2}{*}{#1}}
\newcommand{\RomanNumeralCaps}[1]
    {\MakeUppercase{\romannumeral #1}}
\usepackage[breaklinks=true,bookmarks=false]{hyperref}

\begin{document}
\pagestyle{headings}
\mainmatter
\def\ECCVSubNumber{1746}  

\title{Blind Image Decomposition} 


\titlerunning{Blind Image Decomposition}
\authorrunning{J. Han et al.}

\author{Junlin Han\inst{1,2} Weihao Li\inst{1} Pengfei Fang\inst{1,2} Chunyi Sun\inst{2} Jie Hong\inst{1,2}  Mohammad Ali Armin\inst{1}  Lars Petersson\inst{1}  Hongdong Li\inst{2}}

\institute{Data61-CSIRO\inst{1} 
\; \; 
Australian National University\inst{2} \\
}
\maketitle

\begin{abstract}
We propose and study a novel task named \textbf{B}lind \textbf{I}mage \textbf{D}ecomposition (BID), which requires separating a superimposed image into constituent underlying images in a blind setting, that is, both the source components involved in mixing as well as the mixing mechanism are unknown. For example, rain may consist of multiple components, such as rain streaks, raindrops, snow, and haze. Rainy images can be treated as an arbitrary combination of these components, some of them or all of them. How to decompose superimposed images, like rainy images, into distinct source components is a crucial step toward real-world vision systems. To facilitate research on this new task, we construct multiple benchmark datasets, including mixed image decomposition across multiple domains, real-scenario deraining, and joint shadow/reflection/watermark removal. Moreover, we propose a simple yet general \textbf{B}lind \textbf{I}mage \textbf{De}composition \textbf{N}etwork (BIDeN) to serve as a strong baseline for future work. Experimental results demonstrate the tenability of our benchmarks and the effectiveness of BIDeN. 

Codes and datasets are available at \textcolor{red}{\href{https://github.com/JunlinHan/BID}{GitHub}}.
\end{abstract}

\section{Introduction}
\label{sec:intro}
\begin{figure}[!htbp]
  \begin{minipage}[t]{0.32\linewidth} 
    \centering 
    \includegraphics[width=1.5in, height=0.75in]{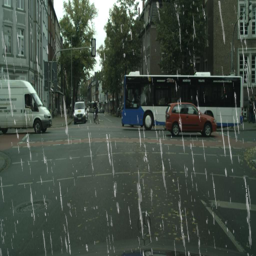}
    \scriptsize \text{(1) rs}
  \end{minipage} 
    \begin{minipage}[t]{0.32\linewidth} 
    \centering 
    \includegraphics[width=1.5in, height=0.75in]{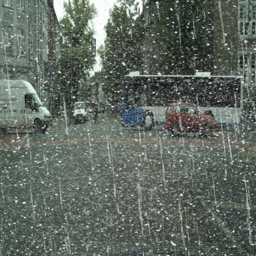}
    \scriptsize \text{(2) rs + snow}
  \end{minipage} 
    \begin{minipage}[t]{0.32\linewidth} 
    \centering 
    \includegraphics[width=1.5in, height=0.75in]{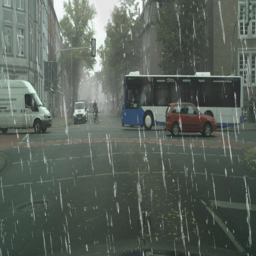}
    \scriptsize \text{(3) rs + light haze }
  \end{minipage} 
  \\
    \begin{minipage}[t]{0.32\linewidth} 
    \centering 
        \includegraphics[width=1.5in, height=0.75in]{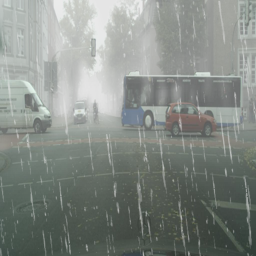}
    \scriptsize \text{(4) rs + heavy haze}
  \end{minipage} 
      \begin{minipage}[t]{0.32\linewidth} 
    \centering 
    \includegraphics[width=1.5in, height=0.75in]{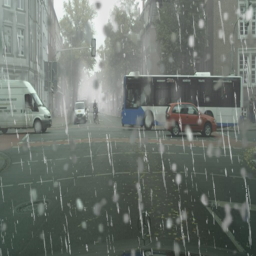}
    \scriptsize \text{(5) rs + mh + raindrop }
  \end{minipage} 
    \begin{minipage}[t]{0.32\linewidth} 
    \centering 
        \includegraphics[width=1.5in, height=0.75in]{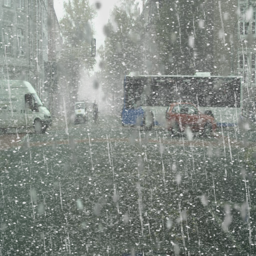}
   \scriptsize \text{(6) rs + snow + mh + raindrop }
  \end{minipage}
  \centering
  \scriptsize \text{rs: rain streak, mh: moderate haze }
  \caption{Example of raining cases. Rain exists in different formats such as rain streak and raindrop. Snow and haze often co-occur during raining. BID setting treats rainy images as an arbitrary combination of these components. Deraining under the BID setting becomes more challenging yet more overarching}
  \label{fig:demo}
\end{figure}

Various computer vision and computer graphics tasks~\cite{gandelsman2019double,zou2020deep,kang2011automatic,le2019shadow,gu2017joint,jayaram2020source,halperin2019neural,alhaija2020intrinsic,fadili2009image,alayrac2019visual} can be viewed as image decomposition, which aims to separate a superimposed image into distinct components/layers with only a single observation. For example, foreground-background segmentation~\cite{faktor2013co,rother2004grabcut,li2018deep,alpert2011image,lin2021real} aims at decomposing a holistic image into foreground objects and background stuff. Image dehazing~\cite{he2010single,berman2016non,li2018single} can be treated as decomposing a hazy image into a haze-free image and a haze map (medium transmission map, atmosphere light). Shadow removal~\cite{le2019shadow,ding2019argan,finlayson2009entropy,jin2021dc,chen2021canet,jin2021dc} decomposes a shadow image into a shadow-free image and a shadow mask. Other tasks like transparency separation~\cite{zhang2018single,fan2017generic,li2020single,ma2019learning}, watermark removal~\cite{liu2021wdnet,chen2021refit}, image deraining~\cite{qian2018attentive,yangjoint2017,zhang2019image,zhou2021image,li2018recurrent,xiao2021improving}, texture separation~\cite{gu2017joint}, underwater image restoration~\cite{galdran2015automatic,han2021underwater}, image desnowing~\cite{liu2018desnownet,ren2017video}, stereo mixture decomposition~\cite{zhong2018stereo}, 3D intrinsic mixture image decomposition~\cite{alhaija2020intrinsic}, fence removal~\cite{du2018accurate,xue2015computational,Liu-CVPR-2020}, flare removal~\cite{wu2021train,asha2019auto} are covered in image decomposition.

Vanilla image decomposition tasks come with a fixed and known number of source components, and the number is most often set to two~\cite{gandelsman2019double,zou2020deep,kang2011automatic,le2019shadow,gu2017joint,zhang2019image,kong2019single}. Such a setting does capture some basic real-world cases. However, real-world scenarios are more complex. Consider autonomous driving on rainy days, where the visual perception quality is degraded by different forms of precipitation and the co-occurring components, shown in Figure~\ref{fig:demo}. Some natural questions emerge: Can a vision system assume precipitations to be of a specific form? Should a vision system assume raindrops always exist or not? Shall a vision system assume the haze or snow comes along with rain or not? These questions are particularly important based on their relevance in real-world applications. The answer to these questions should be \textit{NO}. A comprehensive vision system is supposed to be robust with the ability to handle many possible circumstances~\cite{nayar1999vision,tan2008visibility,li2020all}. Yet, with the previous setting in deraining~\cite{zhang2020beyond,zhou2021pid,qian2018attentive,yangjoint2017,li2019single,li2018recurrent,zhang2021dual,quan2021removing,fu2021unfolding,zhou2021control,zhang2022beyond,zhang2022enhanced} there remains a gap toward sophisticated real-world scenarios. 

This paper aims at addressing the aforementioned gap, as a step toward robust real-world vision systems. We propose a task that: (1) does not fix the number of source components, (2) considers the presence and varying intensities of source components, and (3) amalgamates every source component as potential combinations. To disambiguate with previous tasks, we refer to our proposed task as {\it Blind Image Decomposition (BID)}. This name is inspired by the Blind Source Separation (BSS) task in the field of signal processing. 

The task format is straightforward. We no longer set the number of source components to a fixed value. Instead, we set a maximum number of potential source components, where each component can be arbitrarily part of the mix. Let `A', `B', `C', `D', `E' denote five source components and `a', `b', `c', `d', `e' denote images from the corresponding source components. The mixed image can be either `a', `d', `ab', `bc', `abd', `ade', `acde', `abcde' $\cdots$, with up to 31 possible combinations in total. Given any of the 31 possible combinations as input, a BID method is required to predict and reconstruct the individual source components involved in mixing.

As different components can be arbitrarily involved in the mixing, no existing datasets support such a training protocol. Thus, we construct three benchmark datasets, they are: (I) mixed image decomposition across multiple domains, (II) real-scenario deraining, and (III) joint shadow/reflection/watermark removal. 

To perform multiple BID tasks, we design a simple yet flexible model, dubbed BIDeN (Blind Image Decomposition Network). BIDeN is a generic model that supports diverse BID tasks with distinct objectives. BIDeN is based on the framework of GANs (Generative Adversarial Networks)~\cite{goodfellow2014generative}, and we explore some critical design choices of BIDeN to present a general model. Designed for a more challenging BID setting, BIDeN still outperforms the current state-of-the-art image decomposition model~\cite{zou2020deep,gandelsman2019double} and shows competitive results compared to models designed for specific tasks. Lastly, a comprehensive ablation study is conducted to analyze the design choices of BIDeN.

\section{Related Work}
\label{sec:related}
\noindent \textbf{Image Decomposition.} 
This task is a general task covering numerous computer vision and computer graphics tasks. Double-DIP~\cite{gandelsman2019double} couples multiple DIPs~\cite{ulyanov2018deep} to decompose images into their basic components in an unsupervised manner. Deep Generative Priors~\cite{jayaram2020source} employs a likelihood-based generative model as a prior, performing the image decomposition task. Deep Adversarial Decomposition (DAD)~\cite{zou2020deep} proposed a unified framework for image decomposition by employing three discriminators. A crossroad L1 loss is introduced to support pixel-wise supervision when domain information is unknown. Different from the conventional image decomposition task that usually aims to solve a particular degradation, in the BID setting, we treat degraded images as an arbitrary combination of individual components, and aim to solve sophisticated compound degradation in a unified framework.

\noindent \textbf{Blind Source Separation.} BSS~\cite{hyvarinen2000independent,cichocki2002adaptive,gai2008blindly,gai2009blind,hyvarinen1997fast}, also known as the “cocktail party problem”, is an important research topic in the field of signal processing. Blind refers to a setting where sources and mixing mechanism are both unknown. The task requires performing the separation of a mixture signal into the constituent underlying source signals with limited prior information. A representative algorithm is the Independent Component Analysis (ICA)~\cite{hyvarinen2000independent,lee2000ica,oliveira2008improvements}. The BID task shares common properties with BSS, where the blind settings are similar but not identical. The setting of BID assumes an unknown number of source components involved in the mixing and unknown mixing mechanisms. With such a setting, the number of source components involved in mixing is also unknown. However, the domain information of each source component is considered known as the goal of BID is to advance real-world vision systems. The setting of including known domain information can be better applied to computer vision tasks. For instance, the goal of the shadow removal task is to separate a shadow image into a shadow-free image and a shadow mask, where the domain information is clear.

\begin{figure*}[!htb]
     \centering
     \includegraphics[width = 12cm]
     {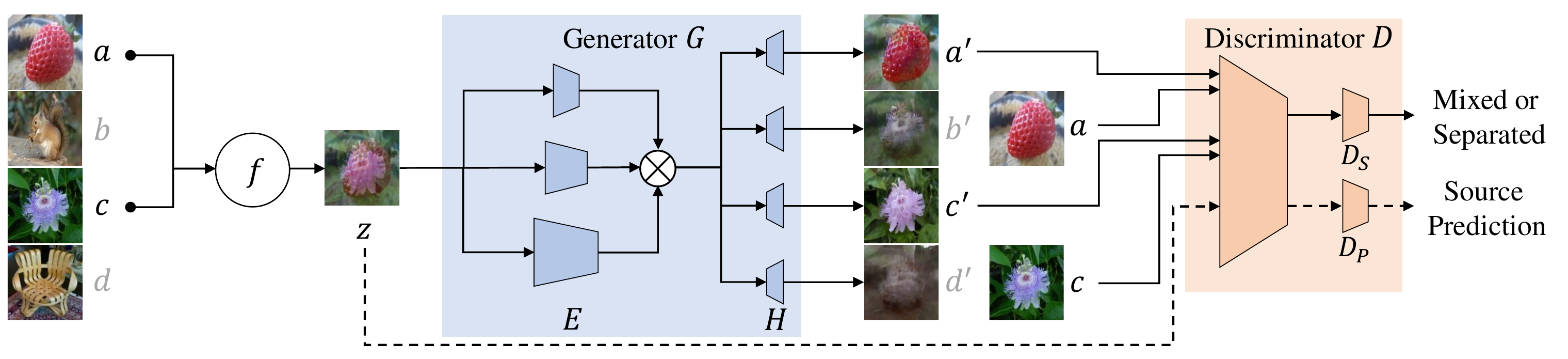}
     \caption{The architecture of the Blind Image Decomposition Network (BIDeN). We show an example, where $N = 4$, $L = 2$, $x = \left\{a,b,c,d \right\}$, and $I = \left\{1,3 \right\}$. $a,c$ are selected then passed to the mixing function $f$, and outputs the mixed input image $z$, which is $f(a,c)$ here. The generator consists of an encoder $E$ with three branches and multiple heads $H$. $\bigotimes$ denotes the concatenation operation. Depth and receptive field of each branch are different to capture multiple scales of features. Each specified head points to the corresponding source component, and the number of heads varies with the maximum number of source components $N$. All reconstructed images ($a',c'$) and their corresponding real images ($a,c$) are sent to an unconditional discriminator. The discriminator also predicts the source components involved in the mixing of the input image $z$. The outputs from other heads ($b',d'$) do not contribute to the optimization}
     \label{fig:net}
\end{figure*}

\noindent \textbf{Generative Adversarial Networks.}
GANs~\cite{goodfellow2014generative} include two key components, a generator, and a discriminator, where the generator is trying to generate realistic samples while the discriminator is trying to identify real samples and generated samples. The adversarial training mechanism helps the output from the generator match the distribution of real data. GANs are especially successful in image generation tasks~\cite{zhang2019self,karras2019style} and image-to-image translation tasks~\cite{isola2017image,han2021dcl,zhu2017unpaired}. GANs are also a common tool for image decomposition tasks, where GANs have been successfully employed in image deraining~\cite{zhang2019image,qian2018attentive}, transparency separation~\cite{ma2019learning,zhang2018single}, and image dehazing~\cite{li2018single}.

\section{Blind Image Decomposition Formulation}
\label{sec:format}
Given a set of $N$ ($N \geq 2$) source components, \ie{}, image domains, denoted by ${\mathcal{X}} = \{X_m\}_{m=1}^N$. Each source component $X_{m}$ contains some images $x_{m}$, $ x_{m} \in X_{m} $.
$L$ ($1 \leq L \leq N$) source components are randomly selected from $\mathcal{X}$. Let $I = \{I_j\}_{j=1}^L$ indicate the index set for the selected source components, where $I_j \in \left\{1,...,N \right\}$. Hence, the selected source components are denoted by
$\{X_{I_j}\}_{j=1}^L$. Each selected source component $X_{I_j}$ contains some images $x_{I_j}$.
With a predetermined mixing function $f$, the mixed image ${z}$ is given by ${z} = f(\{x_{I_j}\}_{j=1}^L)$. The mixed image $z$ can be identical to a single image when $L = 1$. The BID task requires the BID method to find a function $g$ to separate ${z}$, as $g(z) = \{x'_{I_j}\}_{j=1}^L$. Each reconstructed image $x'_{I_j}$ is close to its corresponding image $x_{I_j}$. That is, given a mixed image as input, the task requires the BID method to: (1) predict the source components involved in mixing, (2) reconstruct images preserving the fidelity of the corresponding images involved in mixing.

For the example shown in Figure~\ref{fig:net}, where $N = 4$, $\mathcal{X} = \left\{A,B,C,D\right\}$, $x = \left\{a,b,c,d \right\}$, $L$ can be $1,2,3,4$, $z$ can be $a, b, c, d, f(a,b), f(a,c), f(a,d), f(b,c), f(b,d),\\
f(c,d), f(a,b,c), f(a,b,d), f(a,c,d), f(b,c,d), f(a,b,c,d)$. Let $L = 2$ and $I = \left\{1,3 \right\}$. Given $z = f(a,c) $ as the input, knowing there are four different source components $A,B,C,D$ without any other information, the task requires the method to find a function $g$, so that $g(f(a,c)) = a',c'$, where $a' \to a$ and $c' \to c$. Also, the method should correctly predict the source components involved in the mixing, that is, predicting $I = \left\{1,3 \right\}$.

The BID task is challenging for the following reasons: (1) When $N$ increases, the number of possible $z$ increases rapidly. The BID setting forces the method to deal with $2^{N}-1$ possible combinations. For instance, when $N$ increases to 8, there are 255 variants of $z$. (2) The task requires the method to predict the source components involved in the mixing. Source components are difficult to be predicted when $N$ is large and $L$ varies a lot. (3) The mixing mechanism, or the mixing function $f$, is unknown to the method. The mixing function $f$ varies with different source components and can be non-linear/complex in specific circumstances, such as rendering raindrop images, adding shadows or reflections to images. (4) As $L$ increases, each source component contributes a decreasing amount of information to $z$, making the task highly ill-posed.

\section{Blind Image Decomposition Network}
To perform diverse BID tasks, a unified framework is required. Inspired by the success of image-to-image translation models~\cite{isola2017image,han2021dcl,zhu2017unpaired}, 
we design our Blind Image Decomposition Network (BIDeN) as follows. Figure~\ref{fig:net} presents an overall architecture of BIDeN when the maximum number of source components is four.

The generator $G$ consists of two parts: a multi-scale encoder $E$ and multiple heads $H$. We design a multi-scale encoder containing three branches to capture multiple scales of features. This design is beneficial to the reconstruction of source components. We concatenate different scales of features and send them to multiple heads, where the number of heads is identical to the maximum number of source components. Each head is specific to reconstructing a particular kind of source component. Such multiple-head domain-specific autoencoders have been adopted particularly in inverse rendering task~\cite{halperin2019neural,wang2021learning}. 

The discriminator $D$ consists of two branches and most weights are shared. The reconstructed images and corresponding real images are sent to the discriminator branch $D_{S}$ (Separation) individually. The function of $D_{S}$ is similar to a typical discriminator, \ie{}, classifying whether the input to $D_{S}$ is generated or real to direct the generator generating realistic images. The discriminator branch $D_{P}$ (Prediction) predicts the source components involved in the mixed image $z$, with a confidence threshold of zero. A successful prediction unveils the correct index set of the selected source components $I$. 

Taking the example of Figure~\ref{fig:net}, we name four heads as $\text{H}_{A}$, $\text{H}_{B}$, $\text{H}_{C}$, and $\text{H}_{D}$. For an input $f(a,c)$, $\text{H}_{A}$ and $\text{H}_{C}$ aim to reconstruct $a',c'$, so that $a' \to a$ and $c' \to c$ while $\text{H}_{B}$, $\text{H}_{D}$ are free to output anything or are simply turned off. We employ adversarial loss~\cite{goodfellow2014generative}, perceptual loss~\cite{johnson2016perceptual}, L1/L2 loss, and binary cross-entropy loss. The details of the objective function are expressed below.

\subsection{Objective}
We employ the adversarial loss~\cite{goodfellow2014generative} to encourage the generator $G$ to output well separated and realistic images, regardless of the source components. For the function $g(z) = \{x'_{I_j}\}_{j=1}^L$, the GAN loss is expressed as:
\begin{equation}
\begin{aligned}
\mathcal{L}_{\mathrm{GAN}}\left(G, D_{S}\right) =& \mathbb{E}_{x}\left[\log D_{S}(x)\right] \\
                                                 &+\mathbb{E}_{z}\left[\log \left(1-D_{S}(G(z))\right]\right.,
\end{aligned}
\end{equation}
where $G$ behaves as $g$. It tries to separate the input mixed image $z$ and reconstruct separated outputs $x'_{I_j}$, while $D_{S}$ attempts to distinguish between $G(z)$ and real samples $x_{I_j}$. Note $x$ inside equations 1-5 denotes $x_{I_j}$. We employ the LSGAN~\cite{mao2017least} loss and the Markovian discriminator~\cite{isola2017image}.

The reconstructed images $x'_{I_j}$ should be separated, as well as to be near the corresponding $x_{I_j}$ in a distance sense. Hence, we employ perceptual loss (VGG loss)~\cite{johnson2016perceptual} and L1/L2 loss. They are formalized as:
\begin{equation} 
\begin{array}{l}
\mathcal{L}_{\text{VGG}}(G) = \mathbb{E}_{x,z}[\sum_{l} \lambda_{l}\left[\|\Phi_{l}(x)-\Phi_{l}(G(z)))\|_{1}\right],
\end{array}
\end{equation}
\begin{equation}
    \mathcal{L}_{\text{L1}}(G) = \mathbb{E}_{x,z} \left[\|x-G(z)\|_{1}\right],
\end{equation}
\begin{equation}
    \mathcal{L}_{\text{L2}}(G) = \mathbb{E}_{x,z}\left[\|x-G(z)\|_{2}^{2}\right],
\end{equation}
where $\Phi$ is a trained VGG19~\cite{simonyan2014very} network, $\Phi_{l}$ denotes a specific layer, and $\lambda_{l}$ denotes the weights for the $l$-th layer. The choice of layers and weights is identical to pix2pixHD~\cite{wang2018stacked}. We use L2 loss for masks and L1 loss for other source components. For simplification, we denote L1/L2 loss as $\mathcal{L}_{\text{L}}$.

For the source prediction task, we find that the discriminator performs better than the generator. The goal of the discriminator is to classify between reconstructed samples and real samples. It naturally learns an embedding. Such an embedding is beneficial even when the input is a mixed image $z$. The discriminator $D$ is capable of performing an additional source prediction task. Thus we design a source prediction branch $D_{P}$. The binary cross-entropy loss is employed for the source prediction task:
\begin{equation}
\begin{aligned}
\mathcal{L}_{\text{BCE}}(D_{P}) = &\mathbb{E}_{z} [-\sum^{N}_{m=1}    [GT(z)_{m}\log (D_{P}(z)_{m}) \\
                                  & +(1-GT(z)_{m}) \log (1-D_{P}(z)_{m})]],
\end{aligned}
\end{equation}
where $N$ denotes the maximum number of source components, $GT$ denotes the binary label of the source components involved in the mixing of input image $z$. $D_{P}$ is the source prediction branch of the discriminator. 

Our final objective function is:
\begin{equation}
\begin{aligned}
\mathcal{L}(G,D_{S}, D_{P})=& \lambda_{\text{GAN}}\mathcal{L}_{\mathrm{GAN}}\left(G, D_{S}\right) + \lambda_{\text{VGG}} \mathcal{L}_{\text{VGG}}(G) 
\\
&+\lambda_{\text{L}} \mathcal{L}_{\text{L}}(G) + \lambda_{\text{BCE}} \mathcal{L}_{\text{BCE}}(D_{P}).
\end{aligned}
\end{equation}
We set $\lambda_{\text{GAN}}, \lambda_{\text{VGG}}, \lambda_{\text{L}}, \lambda_{\text{BCE}}$ to be 1, 10, 30, 1 respectively. This setting is a generic setting that is applied to all tasks.

\subsection{Training details}
Throughout all experiments, we use the Adam optimizer~\cite{kingma2014adam} with $\beta_{1}$ = 0.5 and $\beta_{2}$ = 0.999 for both $G, D$. BIDeN is trained for 200 epochs with a learning rate of 0.0003. The learning rate starts to decay linearly after half of the total epochs. We use a batch size of 1 and instance normalization~\cite{ulyanov2016instance}. All training images are loaded as $286 \times 286$ then cropped to $256 \times 256$ patches. Horizontal flip is randomly applied. At test time, we load test images in a $256 \times 256$ resolution. More details on the training settings, the architecture, the number of parameters, and training speed are provided in the Appendix~\ref{appendix: implementation}. The training details of all baselines are also provided there.

\section{Blind Image Decomposition Tasks}
\label{sec:dataset}
We construct benchmark datasets from different views to support practical usages of BID. For each benchmark dataset, we involve multiple source components that may occur together.
To explore the generality of BIDeN and the tenability of constructed datasets, we test BIDeN on three new challenging datasets. BIDeN is trained under the BID setting, which is more difficult than the conventional image decomposition setting. During training, mixed images are randomly synthesized. At test time, the input mixed images are fixed. As BID is a novel task not previously investigated, no existing baselines are available for comparison. For different tasks, we choose different evaluation strategies and baselines.

Throughout all tasks, BIDeN is trained under the BID setting, that is, BIDeN is facing more challenging requirements than other baselines. Also, BIDeN is a generic model designed to perform all kinds of BID tasks. These two constraints limit the performance of BIDeN. We compare BIDeN to other baselines designed for specific tasks, where BIDeN is still able to show very competitive results on all tasks. All qualitative results are \textbf{\textit{randomly picked}}. Additional task settings, discussion on the order of mixing, dataset construction details, results, including the detailed case results of BIDeN, are provided in the Appendix~\ref{appendix: tasks}.

\subsection{Task I: Mixed image decomposition across multiple domains}
\begin{figure*}[!htb]
     \centering
     \includegraphics[width = 12cm]
     {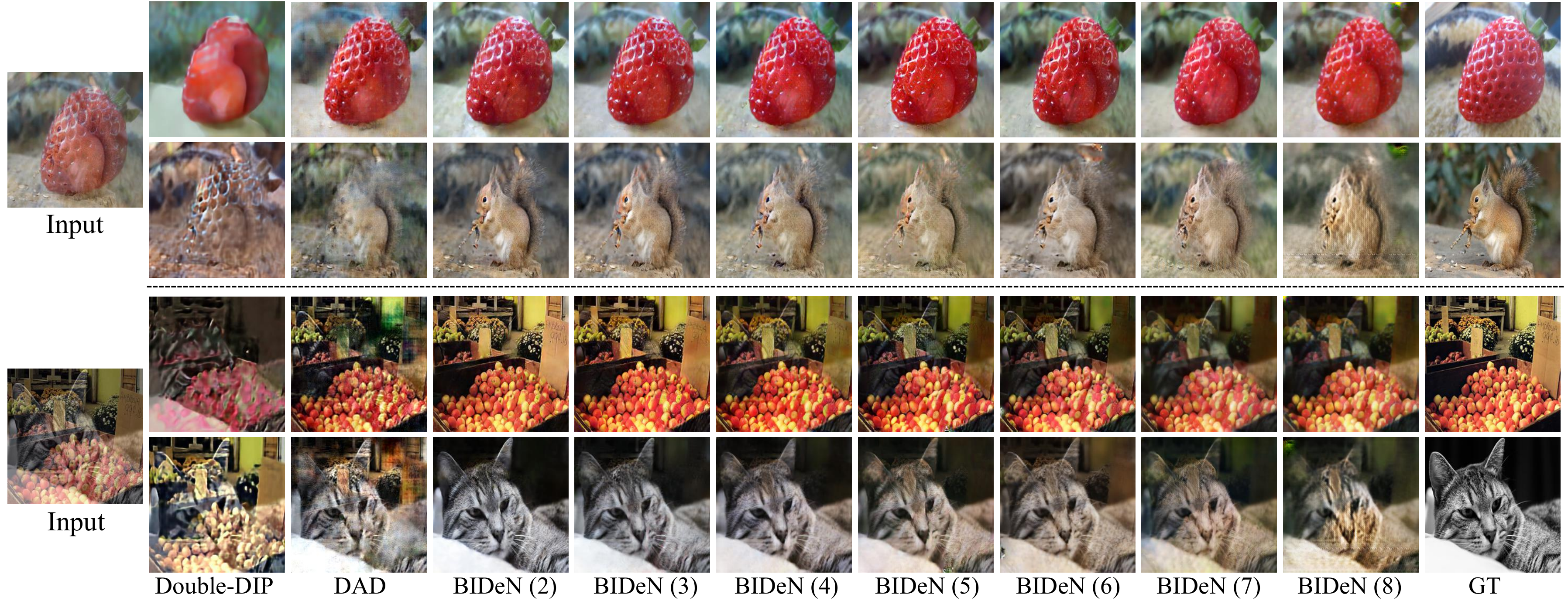}
     \caption{Qualitative results of Task I (Mixed image decomposition across multiple domains). We train BIDeN 7 times, setting different maximum numbers of source components (2-8). Double-DIP fails to separate the mixed input. DAD shows blurry, non-clean results while the results shown by BIDeN are well-separated and visually satisfying. When BIDeN is trained on a greater maximum number of source components, the quality of the results drops progressively as expected}
     \label{fig:res1}
\end{figure*}
\noindent \textbf{Dataset.} This dataset contains eight different image domains, \ie{}, source components. Each domain has approximately 2500 to 3000 images in the training set, and the test set contains 300 images for each domain. Image domains are designed to be big and inclusive to cover multiple categories, like animal, fruit, vehicle, instead of being comparatively small domains such as horse, cat, car. The eight domains are Fruit (2653), Animal (2653), Flower (2950), Furniture (2582), Yosemite (2855), Vehicle (2670), Vegetable (2595), and CityScape (2975). CityScape and Flower are selected from the CityScape~\cite{cordts2016cityscapes} dataset and the VGG flower~\cite{nilsback2006visual} dataset. The remaining six image domains are mainly gathered from Flicker using the corresponding keyword, except for Yosemite, which also combines the Summer2Winter dataset from CycleGAN~\cite{zhu2017unpaired}. The order of the eight domains is randomly shuffled. The mixing mechanism is linear mix.

\begin{table*}[!htbp]
  \centering
  \fontsize{7}{3}\selectfont
       \caption{Quantitative results on Task I (Mixed image decomposition across multiple domains). The testing condition is identical, using Fruit (A) + Animal (B) mixture as inputs. $N$ in BIDeN ($N$) denotes the maximum number of source components. Double-DIP~\cite{gandelsman2019double} performs poorly. Under a more challenging BID setting, BIDeN (2,3,4) still outperforms DAD~\cite{zou2020deep} overall, suggesting the superiority of BIDeN. Please refer to Appendices for  detailed case results}
    \begin{tabular}{lc|c|c|c|c|c|cccc}
    \toprule
     \Rows{Method}&\multicolumn{3}{c|}{Fruit (A)}&\multicolumn{3}{c|}{Animal (B)}& \Rows{Acc (AB) $\uparrow$} & \Rows{Acc (All) $\uparrow$} & \Rows{Model size}\cr
    &PSNR$\uparrow$ & SSIM$\uparrow$ & FID$\downarrow$ & PSNR$\uparrow$ & SSIM$\uparrow$ & FID $\downarrow$\cr
    \midrule
    Double-DIP~\cite{gandelsman2019double}& 13.14&0.49 &257.80 & 13.11&0.39 &221.76 & - & - & - \cr
    DAD~\cite{zou2020deep}& 17.59&0.72 &137.66 & 17.52&0.62 &126.32 & 0.996 & - & 669.0 MB \cr
    \cmidrule(lr){1-10} 
    BIDeN (2)& 20.07&0.79 &62.99 & 19.89&0.69 &69.35 & 1.0 & 0.957 & 144.9 MB \cr
    BIDeN (3)& 19.04&0.75 &74.68 & 18.75 &0.61 &88.23 & 0.836 & 0.807 & 147.1 MB \cr   
    BIDeN (4)& 18.19&0.73 &79.03 & 18.03&0.58 &97.16 & 0.716 & 0.733 & 149.3 MB\cr
    BIDeN (5)& 17.66&0.71 &81.17 & 17.27 &0.54 &114.40 & 0.676 & 0.603 & 151.5 MB \cr 
    BIDeN (6)& 17.28&0.69 &85.64 & 16.57&0.51 &118.00 & 0.646 & 0.483 & 153.7 MB \cr
    BIDeN (7)& 16.70&0.68 &97.26 & 16.54 &0.49 &126.66 & 0.413 & 0.310 &155.9 MB \cr  
    BIDeN (8)& 16.49&0.67 &105.61 & 15.79 &0.45 &191.29 & 0.383 & 0.278 &158.1 MB \cr          
    \bottomrule
    \end{tabular}
     \label{tab:1}
\end{table*}

\noindent \textbf{Experiments and results.} We compare BIDeN to Double-DIP~\cite{gandelsman2019double} and DAD~\cite{zou2020deep}. DAD is trained on the first two domains (Fruit, Animal) with mixed input only. For BIDeN, we train it 7 times under the BID setting, varying from 2 domains to 8 domains. At test time, we evaluate the separation results on Fruit + Animal mixture withPSNR, SSIM~\cite{ssim2014}, and FID~\cite{TTUR2017}. 

Table~\ref{tab:1} and Figure~\ref{fig:res1} present the results of Task I. In terms of PSNR/SSIM, BIDeN outperforms Double-DIP by a large margin and outperforms DAD when $N$ is less than 5. For FID, BIDeN shows preferable results than DAD even when $N = 7$, showing the superiority of BIDeN. 

\subsection{Task II: Real-scenario deraining}

We design two sub-tasks, real-scenario deraining in driving (Task II.A) and real-scenario deraining in general (Task II.B). 

\noindent \textbf{Task II.A: Real scenario deraining in driving} 

\noindent \textbf{Dataset.} Based on the CityScape~\cite{cordts2016cityscapes} dataset, we construct our real-scenario deraining in driving dataset. We use the test set from the original CityScape dataset as our training set (2975), and the validation set from the original CityScape dataset as our test set (500). The test set for all source components contains a fixed number of 500 images. We use three different masks, including rain streak (1620), raindrop (3500), and snow (3500). These masks cover different intensities. For haze, we use the corresponding transmission maps (2975 x 3) with three different intensities acquired from Foggy CityScape~\cite{sakaridis2018semantic}. The masks for rain streak are acquired from Rain100L and Rain100H~\cite{yangjoint2017} while the masks for snow are selected from Snow100K~\cite{liu2018desnownet}. For raindrop masks, we model the droplet shape and property using the metaball model~\cite{blinn1982generalization}. The locations, numbers, and sizes of raindrops are randomly sampled. Paired refraction maps are generated using refractive model~\cite{cohen1998appearance,porav2019can}. The mixing mechanism for this dataset is based on physical imaging models~\cite{yangjoint2017,liu2018desnownet,he2010single,cohen1998appearance,porav2019can}. 
\begin{figure*}[!htb]
     \centering
     \includegraphics[width = 12cm]
     {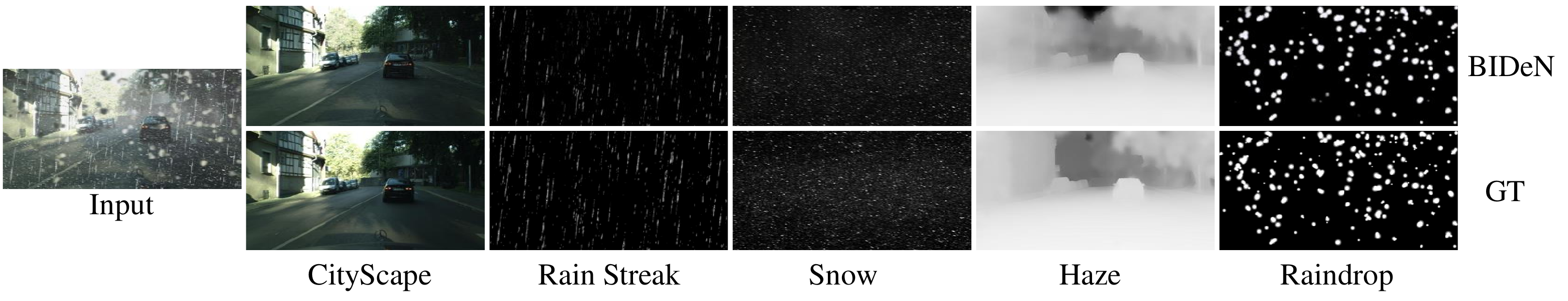}
     \caption{CityScape, masks (Rain Streak, Snow, Raindrop), and transmission map (Haze) generated by BIDeN for case (6), rain streak + snow + moderate haze + raindrop. All generated images are perceptually faithful and visually close to the ground truth (GT)}
     \label{fig:rain_mask}
\end{figure*}
\begin{table*}[!htbp]
  \centering
  \fontsize{7}{3}\selectfont
       \caption{Results of BIDeN on Task II.A (Real-scenario deraining in driving). We employ PSNR and SSIM metrics for both CityScape images, masks, and transmission maps. We report the results for 6 test cases as presented in Figure~\ref{fig:demo}, the 6 cases are (1): rain streak, (2): rain streak + snow, (3): rain streak + light haze, (4): rain streak + heavy haze, (5): rain streak + moderate haze + raindrop, (6) rain streak + snow + moderate haze + raindrop. Note that only haze is divided into light/moderate/heavy intensities. Both training set and test set of Rain Streak, Snow, and Raindrop already consist of different intensities}
    \begin{tabular}{lc|c|c|c|c|c|c|c|c|c|c}
    \toprule
     \Rows{Method}&\multicolumn{2}{c|}{CityScape}&\multicolumn{2}{c|}{Rain Streak}&\multicolumn{2}{c|}{Snow}&\multicolumn{2}{c|}{Haze}&\multicolumn{2}{c|}{Raindrop}& \Rows{Acc $\uparrow$}\cr
    &PSNR$\uparrow$ & SSIM$\uparrow$ & PSNR$\uparrow$ & SSIM$\uparrow$& PSNR$\uparrow$ & SSIM$\uparrow$ & PSNR$\uparrow$ & SSIM$\uparrow$ & PSNR$\uparrow$ & SSIM$\uparrow$\cr
    \midrule
    BIDeN (1)& 30.89&0.932 & 32.13&0.924 & - & - & - & - & - & - & 0.998 \cr
    BIDeN (2)& 29.34&0.899 & 29.24&0.846 & 25.77&0.692 & - & - & - & - & 0.996 \cr    
    BIDeN (3)& 28.62&0.919 & 31.48&0.914 & - & - & 30.77&0.960 & - & -& 0.994 \cr
    BIDeN (4)& 26.77&0.898 & 30.57&0.897 & - & - & 33.73&0.957 & - & -& 0.998 \cr   
    BIDeN (5)& 27.11&0.898 & 30.54&0.898 & - &- & 30.52&0.952 & 20.20&0.908 & 0.994 \cr
    BIDeN (6)& 26.44&0.870 & 28.31&0.823 & 24.79&0.658 & 29.83&0.948 & 21.47&0.893 & 0.998 \cr     
    \bottomrule
    \end{tabular}
     \label{tab:2}
\end{table*}

\noindent \textbf{Experiments and results.} We train BIDeN, MPRNet~\cite{Zamir2021MPRNet}, Restormer~\cite{zamir2021restormer}, and RCDNet~\cite{rcdnet} under the BID setting. For all baselines, we do not require the prediction of the source component and the generation of masks. Thus, BIDeN is still trained with a more challenging requirement. We report the results for 6 cases, as the examples presented in Figure~\ref{fig:demo}. Note that only haze is divided into light/moderate/heavy intensities. Both training set and test set of rain streak, snow, and raindrop already contain different intensities. We report the results in SSIM and PSNR for CityScape images, masks, and transmission maps.
\begin{figure}[!htbp]
  \begin{minipage}[t]{0.16\linewidth} 
    \centering 
    \text{\small Input}
  \end{minipage} 
    \begin{minipage}[t]{0.16\linewidth} 
    \centering 
    \text{\small MPRNet}
  \end{minipage}
      \begin{minipage}[t]{0.16\linewidth} 
    \centering 
    \text{\small Restormer}
  \end{minipage} 
     \begin{minipage}[t]{0.16\linewidth} 
    \centering 
    \text{\small RCDNet}
  \end{minipage}  
    \begin{minipage}[t]{0.16\linewidth} 
    \centering 
      \text{\small BIDeN}
  \end{minipage} 
    \begin{minipage}[t]{0.16\linewidth} 
    \centering 
        \text{\small GT}
  \end{minipage}
  \\
  \begin{minipage}[t]{0.16\linewidth} 
    \centering 
    \includegraphics[width=0.68in, height=0.34in]{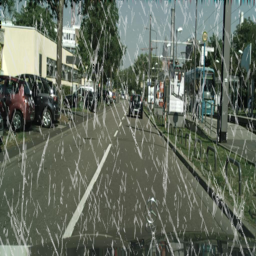}
  \end{minipage} 
    \begin{minipage}[t]{0.16\linewidth} 
    \centering 
        \includegraphics[width=0.68in, height=0.34in]{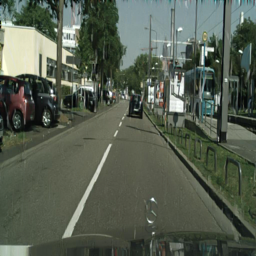}
  \end{minipage}
    \begin{minipage}[t]{0.16\linewidth} 
    \centering 
        \includegraphics[width=0.68in, height=0.34in]{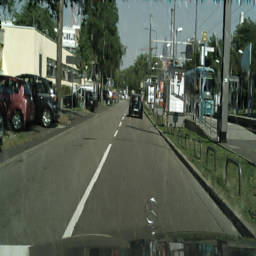}
  \end{minipage}  
      \begin{minipage}[t]{0.16\linewidth} 
    \centering 
        \includegraphics[width=0.68in, height=0.34in]{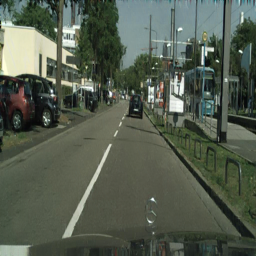}
  \end{minipage}
   \begin{minipage}[t]{0.16\linewidth} 
    \centering 
    \includegraphics[width=0.68in, height=0.34in]{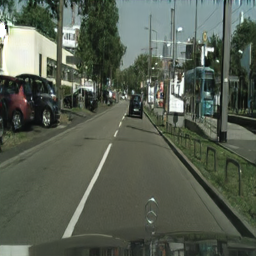}
  \end{minipage} 
    \begin{minipage}[t]{0.16\linewidth} 
    \centering 
        \includegraphics[width=0.68in, height=0.34in]{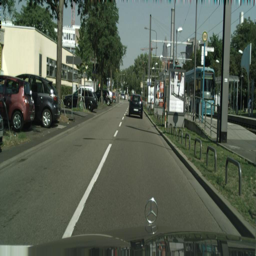}
  \end{minipage} 
\\
  \begin{minipage}[t]{0.16\linewidth} 
    \centering 
    \includegraphics[width=0.68in, height=0.34in]{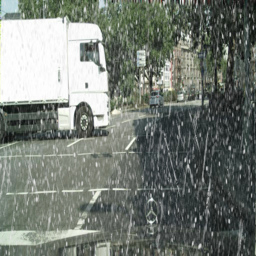}
  \end{minipage} 
    \begin{minipage}[t]{0.16\linewidth} 
    \centering 
        \includegraphics[width=0.68in, height=0.34in]{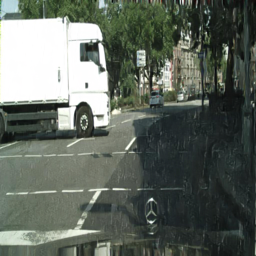}
  \end{minipage}
      \begin{minipage}[t]{0.16\linewidth} 
    \centering 
        \includegraphics[width=0.68in, height=0.34in]{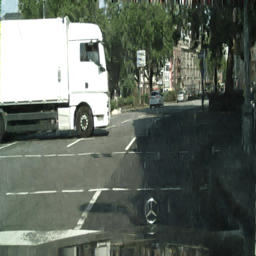}
  \end{minipage}  
      \begin{minipage}[t]{0.16\linewidth} 
    \centering 
        \includegraphics[width=0.68in, height=0.34in]{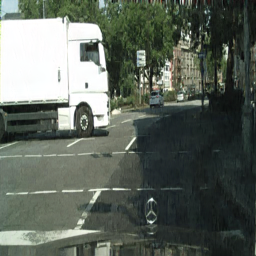}
  \end{minipage}
   \begin{minipage}[t]{0.16\linewidth} 
    \centering 
    \includegraphics[width=0.68in, height=0.34in]{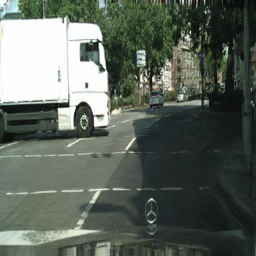}
  \end{minipage} 
    \begin{minipage}[t]{0.16\linewidth} 
    \centering 
        \includegraphics[width=0.68in, height=0.34in]{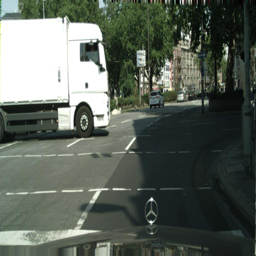}
  \end{minipage} 
\\
  \begin{minipage}[t]{0.16\linewidth} 
    \centering 
    \includegraphics[width=0.68in, height=0.34in]{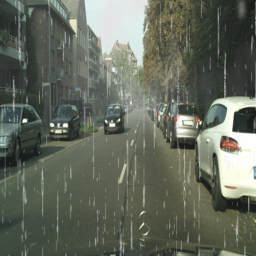}
  \end{minipage} 
    \begin{minipage}[t]{0.16\linewidth} 
    \centering 
        \includegraphics[width=0.68in, height=0.34in]{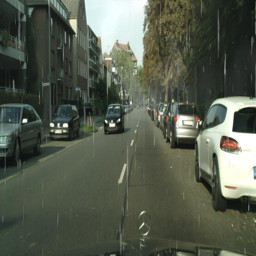}
  \end{minipage}
      \begin{minipage}[t]{0.16\linewidth} 
    \centering 
        \includegraphics[width=0.68in, height=0.34in]{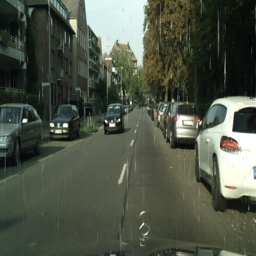}
  \end{minipage}  
      \begin{minipage}[t]{0.16\linewidth} 
    \centering 
        \includegraphics[width=0.68in, height=0.34in]{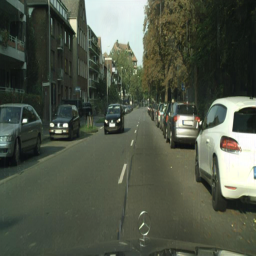}
  \end{minipage}
   \begin{minipage}[t]{0.16\linewidth} 
    \centering 
    \includegraphics[width=0.68in, height=0.34in]{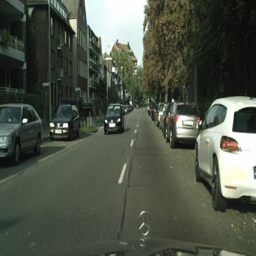}
  \end{minipage} 
    \begin{minipage}[t]{0.16\linewidth} 
    \centering 
        \includegraphics[width=0.68in, height=0.34in]{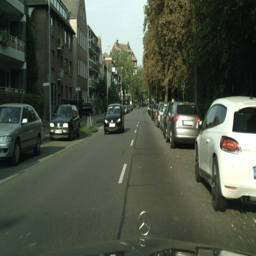}
  \end{minipage} 
\\
  \begin{minipage}[t]{0.16\linewidth} 
    \centering 
    \includegraphics[width=0.68in, height=0.34in]{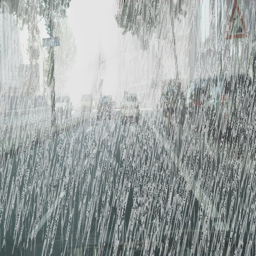}
  \end{minipage} 
    \begin{minipage}[t]{0.16\linewidth} 
    \centering 
        \includegraphics[width=0.68in, height=0.34in]{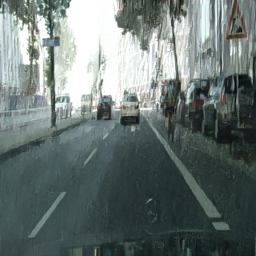}
  \end{minipage}
      \begin{minipage}[t]{0.16\linewidth} 
    \centering 
        \includegraphics[width=0.68in, height=0.34in]{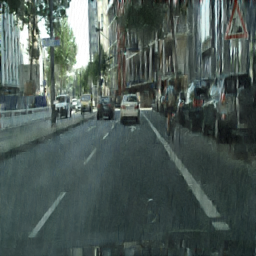}
  \end{minipage}  
      \begin{minipage}[t]{0.16\linewidth} 
    \centering 
        \includegraphics[width=0.68in, height=0.34in]{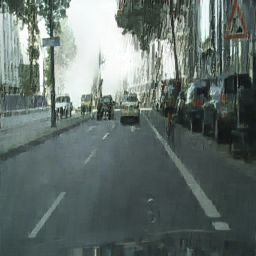}
  \end{minipage}
   \begin{minipage}[t]{0.16\linewidth} 
    \centering 
    \includegraphics[width=0.68in, height=0.34in]{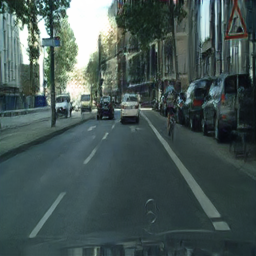}
  \end{minipage} 
    \begin{minipage}[t]{0.16\linewidth} 
    \centering 
        \includegraphics[width=0.68in, height=0.34in]{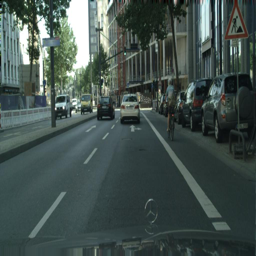}
  \end{minipage} 
\\
  \begin{minipage}[t]{0.16\linewidth} 
    \centering 
    \includegraphics[width=0.68in, height=0.34in]{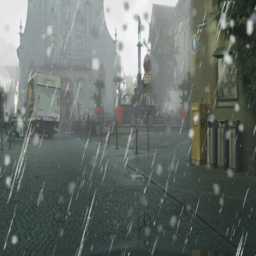}
  \end{minipage} 
    \begin{minipage}[t]{0.16\linewidth} 
    \centering 
        \includegraphics[width=0.68in, height=0.34in]{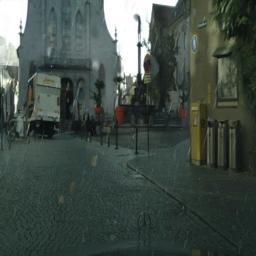}
  \end{minipage}
      \begin{minipage}[t]{0.16\linewidth} 
    \centering 
        \includegraphics[width=0.68in, height=0.34in]{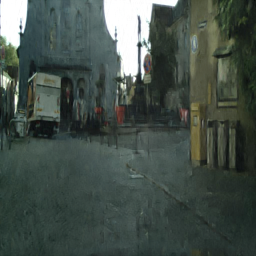}
  \end{minipage}  
      \begin{minipage}[t]{0.16\linewidth} 
    \centering 
        \includegraphics[width=0.68in, height=0.34in]{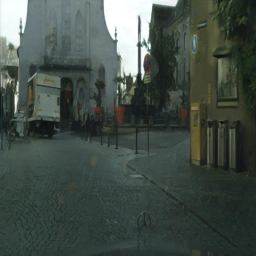}
  \end{minipage}
   \begin{minipage}[t]{0.16\linewidth} 
    \centering 
    \includegraphics[width=0.68in, height=0.34in]{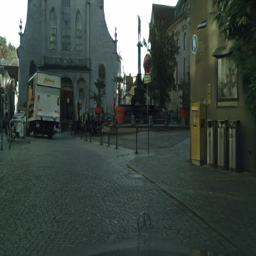}
  \end{minipage} 
    \begin{minipage}[t]{0.16\linewidth} 
    \centering 
        \includegraphics[width=0.68in, height=0.34in]{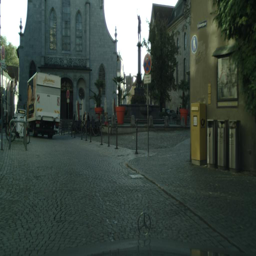}
  \end{minipage} 
\\
  \begin{minipage}[t]{0.16\linewidth} 
    \centering 
    \includegraphics[width=0.68in, height=0.34in]{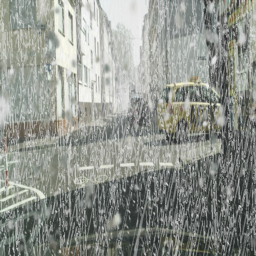}
  \end{minipage} 
    \begin{minipage}[t]{0.16\linewidth} 
    \centering 
        \includegraphics[width=0.68in, height=0.34in]{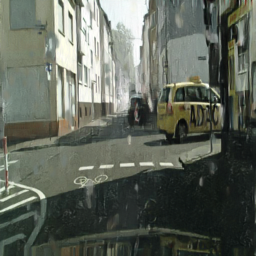}
  \end{minipage}
      \begin{minipage}[t]{0.16\linewidth} 
    \centering 
        \includegraphics[width=0.68in, height=0.34in]{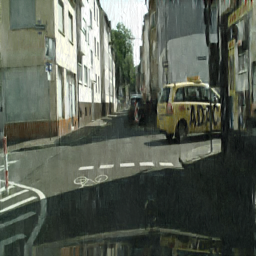}
  \end{minipage}  
      \begin{minipage}[t]{0.16\linewidth} 
    \centering 
        \includegraphics[width=0.68in, height=0.34in]{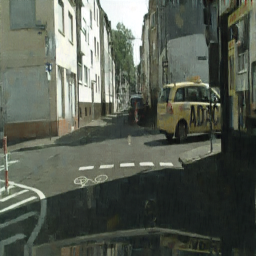}
  \end{minipage}
   \begin{minipage}[t]{0.16\linewidth} 
    \centering 
    \includegraphics[width=0.68in, height=0.34in]{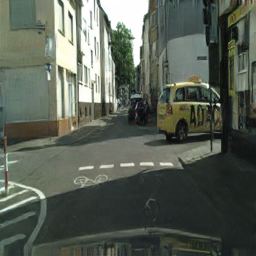}
  \end{minipage} 
    \begin{minipage}[t]{0.16\linewidth} 
    \centering 
        \includegraphics[width=0.68in, height=0.34in]{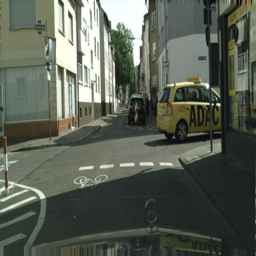}
  \end{minipage} 
  \caption{Results of Task II.A (Real-scenario deraining in driving). Row 1-6 presents 6 cases as presented in Table~\ref{tab:2}. Baselines performs well at case (1) and they effectively removes rain streak but is not strong at removing other components. 
  BIDeN is more robust at the removal of all components. BIDeN generates artifact-free, visually pleasing results while all baselines leaves some components that are not completely removed, especially when hazy intensity is moderate or heavy, as shown in case (4), (5), and (6) }
  \label{fig:rain2}
\end{figure}
\begin{table}[!htbp]
  \centering
  \fontsize{7}{3}\selectfont
       \caption{Comparison on task II.A (Real-scenario deraining in driving) between MPRNet~\cite{Zamir2021MPRNet}, RCDNet~\cite{rcdnet}, and BIDeN. MPRNet and Rostormer shows superior results for case (1) and case (2). In contrast, BIDeN is better at other cases. For the details of 6 test cases, please refer to Table~\ref{tab:2} and Figure~\ref{fig:demo} }
    \begin{tabular}{lc|c|c|c|c|c|c|c|c|c}
    \toprule
     \Rows{Case}& \multicolumn{2}{c|}{Input} & \multicolumn{2}{c|}{MPRNet} & \multicolumn{2}{c|}{Restormer} & \multicolumn{2}{c|}{RCDNet}  & \multicolumn{2}{c}{BIDeN}   \cr
    &PSNR$\uparrow$ & SSIM$\uparrow$ & PSNR$\uparrow$ & SSIM$\uparrow$& PSNR$\uparrow$ & SSIM$\uparrow$& PSNR$\uparrow$ & SSIM$\uparrow$ & PSNR$\uparrow$ & SSIM$\uparrow$ \cr
    \midrule
    \enspace (1)& 25.69&0.786 & 33.39&0.945 & 34.29&0.951  & 32.38&0.937  & 30.89&0.932 \cr
    \enspace (2)& 18.64&0.564 & 30.52&0.909 & 30.60&0.917  & 28.45&0.892 & 29.34&0.899 \cr
    \enspace (3)& 17.45&0.712 & 23.98&0.900 & 23.74&0.905  & 27.14&0.911 & 28.62&0.919 \cr  
    \enspace (4)& 11.12&0.571 & 18.54&0.829 & 20.33&0.853  & 19.67&0.865 & 26.77&0.898 \cr 
    \enspace (5)& 14.05&0.616 & 21.18&0.846 & 22.17&0.859  & 24.23&0.889 & 27.11&0.898 \cr  
    \enspace (6)& 12.38&0.461 & 20.76&0.812 & 21.24&0.821  & 22.93&0.846 & 26.44&0.870 \cr        
    \bottomrule
    \end{tabular}
     \label{tab:22}
\end{table}

For all 6 cases, we report the detailed results of BIDeN in Table~\ref{tab:2}. BIDeN shows excellent quantitative results on accuracy. For the PSNR/SSIM metrics on all source components, BIDeN performs well except for the raindrop masks. An example of all components generated by BIDeN is shown in Figure~\ref{fig:rain_mask}. Table~\ref{tab:22} and Figure~\ref{fig:rain2} presents the comparison between BIDeN and baselines. For better visualization, we resize the resolution of visual examples to match the original CityScape resolution.

\noindent \textbf{Task II.B: Real scenario deraining in general} 

\noindent \textbf{Dataset.}  The training set contains 3661 natural images as rain-free images, where 861 images are adopted from the training set of~\cite{qian2018attentive}, 1800 images are borrowed from the training set of Rain1800~\cite{yangjoint2017}, and the rest 1000 images are selected from the training set of Snow100K~\cite{liu2018desnownet}. We adopt identical rain streak, snow, and raindrop masks from Task II.A. The test set contains real images only, including rain streak (185, from~\cite{li2019single}), raindrop (249, from~\cite{qian2018attentive}) and snow (1329, from~\cite{liu2018desnownet}) images.

\begin{figure}[!htb]
     \centering
     \includegraphics[width = 12cm]
     {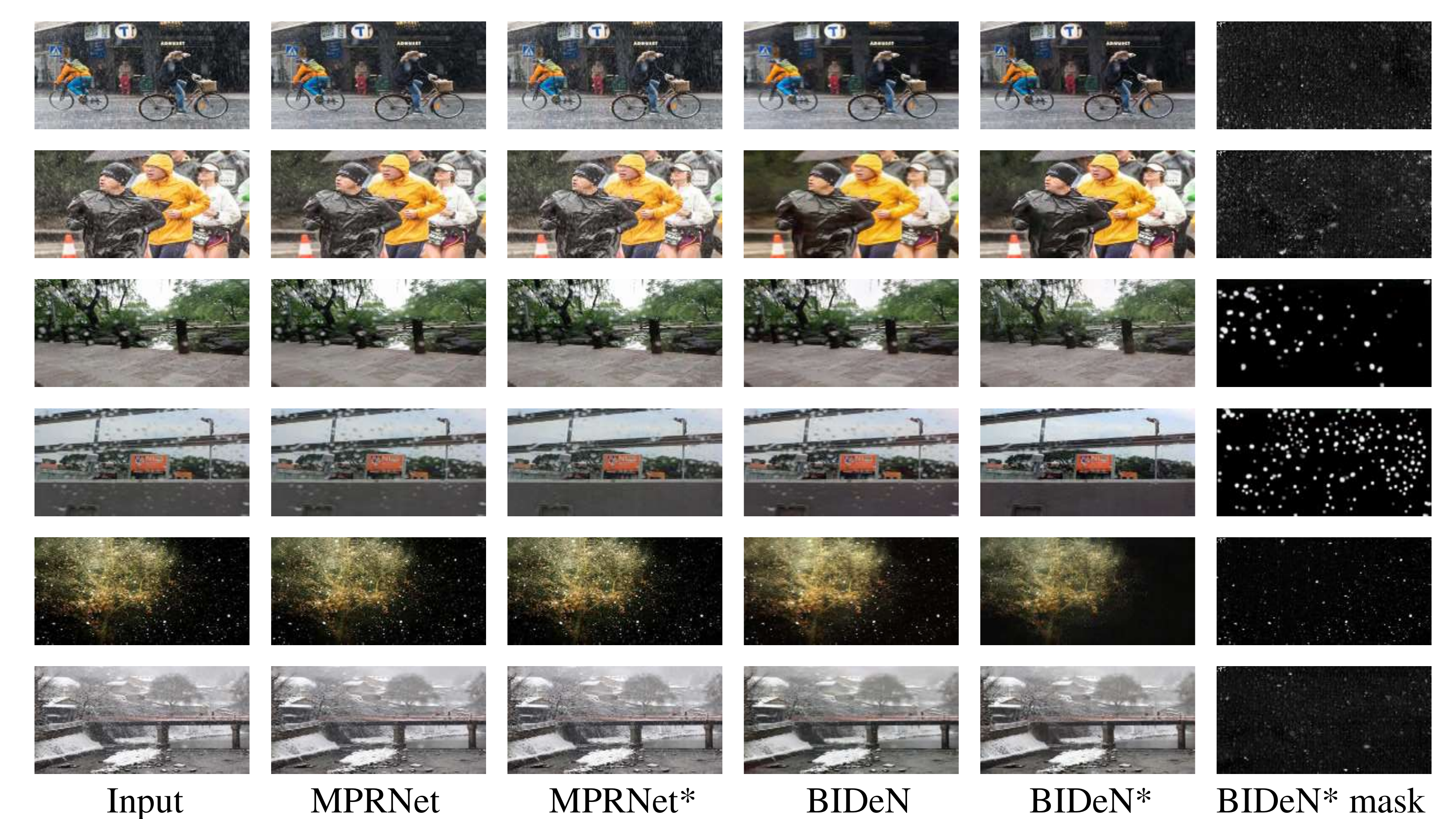}
     \caption{Qualitative results of Task II.B (Real-scenario deraining in general). * indicates the BID setting. Rows 1-2, 3-4, and 5-6 presents the results of real rain streak, raindrop, and snow images, respectively. For all cases, models trained under the BID setting are more robust in restoring real-world images}
     \label{fig:realrain}
\end{figure}

\begin{table}[!htbp]
  \centering
  \fontsize{7}{3}\selectfont
       \caption{ Quantitative results of Task II.B (Real-scenario deraining in general). * indicates the BID setting. For all methods and testing cases, models trained under the BID setting performs better than the conventional image decomposition setting }
    \begin{tabular}{lc|c|c|c|c|c}
    \toprule
     \Rows{Method}& \multicolumn{2}{c|}{Rain Streak} & \multicolumn{2}{c|}{Raindrop} & \multicolumn{2}{c}{Snow}   \cr
    &NIQE$\downarrow$ & BRISQUE$\downarrow$ & NIQE$\downarrow$ & BRISQUE$\downarrow$& NIQE$\downarrow$ & BRISQUE$\downarrow$\cr
    \midrule
    \enspace Input& 4.86&27.84 & 5.61&24.85  & 4.74&22.68 \cr
    \midrule 
    \enspace MPRNet& 4.14&28.72 & 4.94&29.42  & 4.60&25.93 \cr
    \enspace MPRNet*& 4.34&28.00 & 4.81&25.86  & 4.24&24.55 \cr
    \midrule 
    \enspace BIDeN& 4.71&25.39 & 5.39&22.94  & 4.97&22.64 \cr 
    \enspace BIDeN*& 4.31&26.55 & 4.71&21.22 & 4.28&22.40 \cr  
    \bottomrule
    \end{tabular}
     \label{tab:realrain}
\end{table}

\noindent \textbf{Experiments and results.} We aim to validate whether models trained on the synthetic dataset can generalize to real testing samples and whether the BID training setting generalizes better than the conventional image decomposition setting. We train BIDeN and MPRNet under the BID setting as well as the conventional image decomposition setting with identical training recipes, that is, the training recipes used in Task II.A. For the conventional image decomposition setting, models are trained on a large-scale synthetic dataset~\cite{Zamir2021MPRNet} (13712 pairs). We report the quantitative results with no-reference metrics, NIQE~\cite{mittal2012making} and BRISQUE~\cite{mittal2012no}. 

All trained models are evaluated with real-world images. Results are presented in Table~\ref{tab:realrain} and Figure~\ref{fig:realrain}. With only 26.6\% training data and iterations, both BIDeN and MPRNet models trained under the BID setting are more robust in restoring real images, showing the generality and robustness of the BID setting in real-world scenarios.

\subsection{Task III: Joint shadow/reflection/watermark removal}
\begin{figure*}[!htb]
     \centering
     \includegraphics[width = 12cm]
     {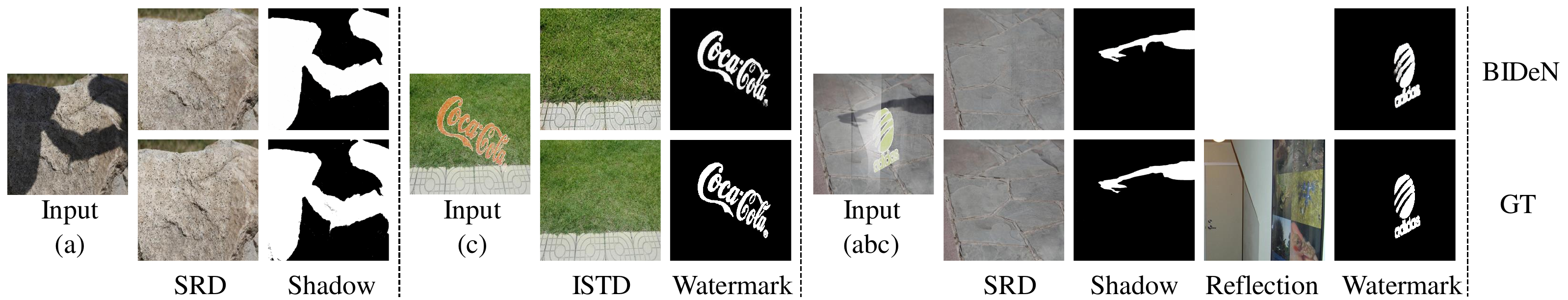}
     \caption{Images and masks (Shadow, Watermark) produced by BIDeN for three cases, (a), (c), and (abc). All generated masks are faithful to the ground truth (GT)}
     \label{fig:shadow_mask}
\end{figure*}

\noindent \textbf{Dataset.} This task is designed to jointly perform multiple tasks with uncertainty in one go. Once a model is trained on this dataset under the BID setting, the model is capable of performing multiple tasks. We construct two versions for this dataset, Version one (V1) is based on ISTD~\cite{wang2018stacked}, and Version two (V2) is based on SRD~\cite{qu2017deshadownet,cun2020towards}. We use paired shadow masks, shadow-free images, and shadow images from ISTD/SRD. ISTD consists of 1330 training images and 540 test images while SRD contains 2680 training images and 408 testing images. The algorithm for adding reflection to images is acquired from~\cite{zhang2018single}, we select 3120 images from the reflection subset~\cite{zhang2018single} as the reflection layer. The watermark generation algorithm as well as the paired watermark masks, RGB watermark images are acquired from LVM~\cite{liu2021wdnet}, we select 3000 paired watermark images and masks from the training set of LVW~\cite{liu2021wdnet}. 

Following the data split of ISTD and SRD, both V1 and V2 share 2580 reflection layer images and 2460 watermark images/masks. V1 contains 1330 paired shadow-free images/shadow masks and the test set contains 540 images for every source component. V2 includes 2680 paired shadow-free images/shadow masks and the test set contains 408 images for every source component. Note that we do not require the reconstruction of reflection layer images.
\begin{table*}[tb]
  \centering
  \fontsize{7}{3}\selectfont
       \caption{Results of Task III (Joint shadow/reflection/watermark removal). We employ RMSE$\downarrow$ to measure shadow region, non-shadow region, and all region. For BIDeN, we report the performance of all cases. a,b,c denotes shadow, reflection, and watermark, respectively. BIDeN (ab) is the result of BIDeN tested on shadow + reflection inputs. Results for all baselines are reported by~\cite{cun2020towards,fu2021auto}. The generality of BIDeN and the challenging BID training setting limit the performance of BIDeN}
    \begin{tabular}{lc|c|c|c|c|c|c|c}
    \toprule
    \Rows{Method} & \multicolumn{4}{c|}{Version one (V1), ISTD} & \multicolumn{4}{c}{Version two (V2), SRD} \cr
    &Shadow&Non-Shadow&All &Acc &Shadow&Non-Shadow&All &Acc \cr
    \midrule
    Bilateral~\cite{yang2012shadow}& 19.82 & 14.83 & 15.63 & - & 23.43 & 22.26 & 22.57 & - \cr
    Regions~\cite{guo2012paired} & 18.95 & 7.46 & 9.30 & - & 29.89 & 6.47 & 12.60 & -\cr    
    Interactive~\cite{gong2014interactive}& 14.98 & 7.29 & 8.53 & - & 19.58 & 4.92 & 8.73 & -\cr 
    DSC~\cite{hu2019direction} & 9.48 & 6.14 & 6.67& - & 10.89 & 4.99 & 6.23& - \cr   
    DHAN~\cite{cun2020towards}& 8.14 & 6.04 & 6.37& - & 8.94 & 4.80 & 5.67& -  \cr 
    Auto-Exposure~\cite{fu2021auto} & 7.77 & 5.56 & 5.92 & - & 8.56 & 5.75 & 6.51 & -\cr
    CANet~\cite{chen2021canet} & 8.86 & 6.07 & 6.15 & - & 7.82 & 5.88 & 5.98 & -\cr   
    \cmidrule(lr){1-9} 
    BIDeN (a)& 11.55 & 10.24 & 10.45 & 0.359 & 12.06 & 7.47 & 8.73 & 0.919\cr
    BIDeN (ab)& 12.96 & 10.77 & 11.12 & 0.661 & 14.10 & 8.16 & 9.79 & 0.911\cr
    BIDeN (ac)& 11.89 & 10.23 & 10.50 & 0.694 & 13.29 & 8.08 & 9.51 & 0.943\cr
    BIDeN (abc)& 13.20 & 10.76 & 11.16 & 0.929 & 15.28 & 8.85 & 10.62 & 0.936\cr
    \cmidrule(lr){1-9}
    BIDeN (b)& - & - & 10.85 & 0.559 & - & - & 8.01 & 0.891\cr
    BIDeN (c)& - & - & 10.20 & 0.461 & - & - & 7.92 & 0.914\cr
    BIDeN (bc)& - & - & 10.77 & 0.727 & - & - & 8.71 & 0.879\cr
    \bottomrule
    \end{tabular}
     \label{tab:3}
\end{table*}

\noindent \textbf{Experiments and results.} We mainly compare the shadow removal results to multiple shadow removal baselines, including~\cite{yang2012shadow,guo2012paired,gong2014interactive,hu2019direction,cun2020towards,fu2021auto,chen2021canet}. We train BIDeN under the BID setting. The trained BIDeN is capable of dealing with all combinations between shadow/reflection/watermark removal tasks. At test time, we report the results for all cases. We employ the root mean square error (RMSE) in LAB color space, following~\cite{cun2020towards,fu2021auto}. 

The quantitative results for Task III are reported in Table~\ref{tab:3}. Constrained by the generality of BIDeN and the challenging BID training setting, BIDeN does not show superior quantitative results compared to other baselines designed for the shadow removal task only. Please refer to the Appendix~\ref{appendix: results} for qualitative results.

\section{Ablation Study and Analysis}
\label{sec:ablation}

We perform ablation experiments to analyze the effectiveness of each component inside BIDeN. Evaluation is performed on Task I (Mixed image decomposition across multiple domains). We set the maximum number of source components to be 4 throughout all ablation experiments. The results are shown in Table~\ref{tab:ablation}.

\begin{table}[!htbp]
  \centering
  \fontsize{7}{3}\selectfont
    \begin{tabular}{lc|c|cc}
    \toprule
     \Rows{Ablation}&\multicolumn{1}{c|}{Fruit (A)}&\multicolumn{1}{c|}{Animal (B)}& \Rows{Acc (AB) $\uparrow$} & \Rows{Acc (All) $\uparrow$}\cr
    &PSNR$\uparrow$  & PSNR$\uparrow$ \cr
    \midrule
    \RomanNumeralCaps{1}& 17.26& 17.05& 0.730 & 0.732 \cr
    \RomanNumeralCaps{2}& 17.95& 17.41& 0.566 & 0.616 \cr
    \RomanNumeralCaps{3}& 16.67& 16.34& 0.706 & 0.722 \cr
    \RomanNumeralCaps{4}& 15.56& 13.65& 0.733 & 0.753 \cr
    \RomanNumeralCaps{5}& 18.04& 17.98& 0.0 & 0.06 \cr
    \RomanNumeralCaps{6}& 18.19& 17.97& 0.634 & 0.698 \cr
    \RomanNumeralCaps{7}& 18.13& 17.98& 0.520 & 0.609 \cr
    \RomanNumeralCaps{8}& 15.68& 15.64& 0.716 & 0.683 \cr
    \cmidrule(lr){1-5} 
    BIDeN & \textbf{18.19}& \textbf{18.03}& \textbf{0.716} & \textbf{0.733 }\cr
    \bottomrule
    \end{tabular}
     \caption{Ablation study on the design choices of BIDeN. (\RomanNumeralCaps{1}) Single-scale encoder. (\RomanNumeralCaps{2}) No adversarial loss. (\RomanNumeralCaps{3}) No perceptual loss. (\RomanNumeralCaps{4}) No L1/L2 loss. (\RomanNumeralCaps{5}) No binary cross-entropy loss. (\RomanNumeralCaps{6}) Source prediction branch inside the generator. (\RomanNumeralCaps{7}) No weights sharing between two branches of discriminator.  (\RomanNumeralCaps{8}) Zeroed loss}
     \label{tab:ablation}
\end{table}

\noindent \textbf{Multi-scale encoder (\RomanNumeralCaps{1})}. We present the results of using a single-scale encoder to replace the multi-scale encoder. BIDeN yields better performance when the multi-scale encoder is employed, which validates the effectiveness of our design.

\noindent \textbf{Choice of losses (\RomanNumeralCaps{2}, \RomanNumeralCaps{3}, \RomanNumeralCaps{4}, \RomanNumeralCaps{5})}. BIDeN consists of four different losses, we show that removing either one of the losses leads to a performance drop. 

\noindent \textbf{Source prediction branch (\RomanNumeralCaps{6}, \RomanNumeralCaps{7})}. We move the prediction branch $D_{P}$ to the generator. This change degrades the performance, showing that the source prediction task is better to be performed by the discriminator.
We report the results for a variant where $D_{P}$ does not share weights with the separation branch $D_{S}$. The performance of this variant is worse than vanilla BIDeN, indicating that the embedding learned by $D_{S}$ is beneficial to $D_{P}$.

\noindent \textbf{Zeroed loss (\RomanNumeralCaps{8})}. Taking the example of Figure~\ref{fig:net}, four heads are $\text{H}_{A}$, $\text{H}_{B}$, $\text{H}_{C}$, and $\text{H}_{D}$. By default, BIDeN ignores the outputs from $\text{H}_{B}$ and $\text{H}_{D}$. Here, we encourage the outputs from $\text{H}_{B}$ and $\text{H}_{D}$ to be zero pixels. Such a zeroed loss forces the generator to perform the source prediction task implicitly, however, the results after applying zeroed loss are not comparable to default BIDeN.

\section{Conclusion}
\label{sec:discussion}
We believe BID is a novel computer vision task advancing real-world vision systems. We form a solid foundation for the future study and we invite the community to further explore its potential, including discovering interesting areas of application, developing novel methods, extending the BID setting, and constructing benchmark datasets. We expect more application areas related to image decomposition, especially in image deraining, to apply the BID setting. 

\clearpage
%
%
\bibliographystyle{splncs04}
\bibliography{egbib}

\clearpage
\appendix
\section{Implementation Details}
\label{appendix: implementation}
\subsection{Training details, running time, and model size}

\noindent \textbf{BIDeN}. We train BIDeN using a Tesla P100-PCIE-16GB GPU. The GPU driver version is 440.64.00 and the CUDA version is 10.2. We initialize weights using Xavier initialization~\cite{glorot2010understanding}. For Task I (Mixed image decomposition across multiple domains), BIDeN (2) to BIDeN (8) takes approximately 37 hours, 50 hours, 61 hours, 71 hours, 82 hours, 91 hours, and 101 hours training time. For Task II.A (Real-scenario deraining in driving), the runtime of BIDeN is approximately 96 hours. However, BIDeN is required to perform additional mask reconstruction and source prediction tasks. By removing additional tasks from the training of BIDeN, the GPU hours can drop to 45 hours. 
 
\noindent \textbf{Double-DIP}. 
We follow the default training setting of Double-DIP~\cite{gandelsman2019double}. We use the official PyTorch implementation (\textcolor{red}{\href{https://github.com/yossigandelsman/DoubleDIP}{link}}). We train a single image for 8000 iterations on a Tesla P100-PCIE-16GB GPU, the GPU driver version is 415.27 and the CUDA version is 10.0. The runtime for a single input image is approximately 20 minutes.

\noindent \textbf{DAD}. 
We follow the default training setting (Epoch 200, batch size 2, image crop size 256) of DAD~\cite{zou2020deep}. Experiments are based on the official PyTorch implementation (\textcolor{red}{\href{https://github.com/jiupinjia/Deep-adversarial-decomposition}{link}}). We train DAD on a Tesla P100-PCIE-16GB GPU. The GPU driver version is 440.64.00 and the CUDA version is 10.2. DAD takes 13 hours of runtime. 

\noindent \textbf{MPRNet}. 
We follow the default training setting (Epoch 250, batch size 16, image crop size 256) of MPRNet~\cite{Zamir2021MPRNet}. For a fair comparison, we apply the same data augmentation operations of BIDeN to MPRNet. We use the official PyTorch implementation (\textcolor{red}{\href{https://github.com/swz30/MPRNet}{link}}) of MPRNet. We train MPRNet using 4 Tesla P100-PCIE-16GB GPU, the GPU driver version is 415.27 and the CUDA version is 10.0. The runtime of MPRNet is 20 hours and the model size of MPRNet is 41.8 MB.

\noindent \textbf{Restormer}. 
We follow the training setting used in \textcolor{red}{\href{https://github.com/leftthomas/Restormer}{link}}. Similar to MPRNet, we apply the same data augmentation operations used in BIDeN to Restormer. We reproduce our results based on a PyTorch implementation (\textcolor{red}{\href{https://github.com/leftthomas/Restormer}{link}}) of Restormer~\cite{zamir2021restormer}. We train Restormer using a GeForce RTX 3900 GPU, the GPU driver version is 510.47 and the CUDA version is 11.6. The runtime of Restormer is approximately 8 hours and the model size is 25.3 MB.

\subsection{Architecture of BIDeN}
Following the naming convention used in CycleGAN~\cite{zhu2017unpaired} and perceptual loss~\cite{johnson2016perceptual}, let \textbf{c3s1-k} denote a 3×3 Convolution-InstanceNorm-ReLU layer with stride 1 and $k$ filters. \textbf{Rk} denotes a residual block that contains two 3×3 convolutional layers with the same number of filters on both layer and \textbf{Rk9} denotes nine continuous residual blocks. \textbf{uk} denotes a 3×3 fractional-strided-Convolution-InstanceNorm-ReLU layer with $k$ filters and stride $\frac{1}{2}$. Let \textbf{Ck} denotes a 4×4 Convolution-InstanceNorm-LeakyReLU (slope 0.2) layer with $k$ filters and stride 2. Both reflection padding and zero padding are employed.

\noindent \textbf{Encoder}.
Our multi-scale encoder $E$ contains three branches, we name them $E_{B1}$, $E_{B2}$ and $E_{B3}$.

$E_{B1}$ consists of c3s2-256, Rk9, c1s1-128. $E_{B2}$ consists of c7s1-64, c3s2-128, c3s2-256, Rk9. $E_{B3}$ contains c15s1-64, c3s2-128, c3s2-256, c3s1-256, c3s1-256, Rk9, c1s1-128. The number of parameters is 33.908 million for the encoder.

\noindent \textbf{Heads}.
The architecture of each head $H$ is: c1s1-256, c1s1-256, u128, u64, c7s1-3.
Each head has 0.575 million parameters.

\noindent \textbf{Discriminator}.
The discriminator $D$ contains two branches, $D_{S}$ (Separation) and $D_{P}$ (Prediction). Most weights are shared, the shared part includes C64, C128, C256. 

The last layer of $D_{S}$ is C512. $D_{P}$ contains c1s1-512 (LeakyReLu with slope 0.2), global max pooling, c1s1-N, where $N$ is the maximum number of source components. The Discriminator has approximately 3.028 million parameters in total. The confidence threshold of $D_{P}$ is 0.

\subsection{Tasks}
\label{appendix: tasks}
\noindent \textbf{Task I: Mixed image decomposition across multiple domains.} 
We use linear mix as the mixing mechanism. We do not introduce additional non-balanced mixing factors or non-linear mixing as Task I is challenging enough.
The mixed image ${z}$ is expressed as 
$z = \frac{1}{L} \sum^{L}_{j=1}  x_{I_j}$. The possibilities of every component to be selected vary with the maximum number of source components $N$. We set the possibilities to be 0.9, 0.8, 0.7, 0.6, 0.5, 0.5, 0.5 for $N =2, 3, 4, 5, 6, 7, 8$, respectively. As the mixing is linear, the order of mixing does not matter.

\noindent \textbf{Task II: Real-scenario deraining.} 
For both Task II.A and Task II.B, the mixing mechanism is based on the physical imaging models~\cite{yangjoint2017,liu2018desnownet,he2010single,cohen1998appearance,porav2019can} and Koschmieder's law. The model for rain streak and snow is:

$I(x) = J(x)(1-m(x)) + A*m(x)$,

and the model for haze is:

$I(x) = J(x)t(x) + A(1-t(x))$,

where $x$ is the pixel of images, $I$ is the observed intensity, $J$ is the scene radiance, $A$ is the global atmospheric light, and $m$ is the mask of rain streak and snow. $t$ denotes the transmission map. We set $A$ between $\left[0.8, 1.0 \right]$ during training, and fix $A = 0.9$ at test time. 

To render the raindrop effect, we define a statistical model to estimate the location and motion of the raindrops. We employ the meta-ball model~\cite{blinn1982generalization} for the interaction effect between multiple raindrops. 

For raindrop positions, we randomly sample it over the entire scene. The raindrop radius is also randomly sampled. A single raindrop is combined with another 1 to 3 smaller raindrops to form a realistic raindrop shape. Each composite raindrop could further merge with other raindrops on the scene. The velocity along the y-axis of the raindrop is proportional to the raindrop radius. Raindrop masks are randomly selected on the time dimension for diversity. 
A simple refractive model~\cite{cohen1998appearance} is employed. We create a look-up table $T$ with three dimensions. The red and green channels together encode the texture of the raindrop, and the blue channel represents the thickness of the raindrops. Then, the texture table $T$ is masked by the alpha mask created by the meta-ball model. The masked table is dubbed $M$. 
The location (x,y) of the world point that is rendered at image location (u, v) on the surface of a raindrop is modeled as: 

$ x =u+(R_{(u,v)}*B_{(u,v)})$,  
$ y =v+(G_{(u,v)}*B_{(u,v)})$.

where $R_{(u,v)}$, $B_{(u,v)}$ denote the pixel at location $(u,v)$ in the red and blue channels of $M$.

We acquire the destination pixel coordinate for location (u,v) based on the above equations and generate the distorted image. We also apply random light reduction and blur to the distorted image. For the reduction, we set the rate $r$ between $\left[0.8, 0.98 \right]$ during training, and fix rate $r = 0.9$ at test time. The reduction can be expressed as $D = rD$, where $D$ is the distorted image and $r$ is the rate.
We use a kernel size of 3 for Gaussian blur. 

At last, we merge the distorted image with the original image:

$I(x) = \alpha(x)O(x) + (1 - \alpha(x)) * D(x)$,

where $x$ denotes the pixel of images, $I$ is the observed intensity, $O$ is the original image, $D$ is the distorted image, $\alpha$ is the value of the raindrop mask generated by the metaball model. 

The probabilities of every component to be selected are 1.0, 0.5, 0.5, 0.5 for rain streak, snow, haze, and raindrop for Task II.A. The mixing order is rain streak, snow, haze, and raindrop. We design this since 1: streak and snow do not change the atmosphere a lot, so they should come first and 2: raindrops usually occur on the top of glass lens and are observed at first, thus they should come last. In Task II.B, the probabilities of every component to be selected are 0.6, 0.5, 0.5 for rain streak, snow, and raindrop. The mixing order is rain streak $\to$ snow $\to$ raindrop.

\noindent \textbf{Task III: Joint shadow/reflection/watermark removal.}
We use paired shadow masks, shadow images, and shadow-free images provided in ISTD~\cite{wang2018stacked} and SRD~\cite{qu2017deshadownet,cun2020towards}. The original SRD does not offer shadow masks, we use the shadow masks generated by Cun \etal{}~\cite{cun2020towards}. 

The algorithm of adding reflection to images~\cite{zhang2018single} is expressed as:

$I(x) = T(x) + R(x) * V(x)$,

where $x$ is the pixel of images, $I$ is the observed intensity, $T$ is the transmission layer, $R$ is the reflection layer, and $V$ denotes the vignette mask. The reflection image $R$ is processed by a Gaussian smoothing kernel with a random kernel size, where the size is in the range of 3 to 17 pixels during training, and fixed to 11 pixels during testing.

For watermarks, we follow the watermark composition model~\cite{liu2021wdnet}. We use the RGB watermark images to add the watermark effect.  We require the BID method to reconstruct the watermark mask. The watermark composition model is:

$I(x) = J(x)(1-w(x)) + A * w(x)$,

where $x$ denotes the pixel of images, $I$ is the observed intensity, $J$ is the scene radiance, $A$ is the global atmospheric light, and $w$ is the watermark image. We set $A$ between $\left[0.8, 1.0 \right]$ during training, and fix $A = 0.9$ for testing. 

The probabilities of every component to be selected are 0.6, 0.5, 0.5 for shadow, reflection, and watermark, respectively. The order of mixing is shadow $\to$ reflection $\to$ watermark, as watermarks are usually added during post-processing

\section{Additional Results}
\subsection{Additional results of Task I}
\noindent \textbf{Detailed case results of BIDeN.} When the maximum number of components $N$ increases, the number of possible $z$ increases rapidly. There are $2^{N}-1$ possible combinations between source components, that is, $2^{N}-1$ cases. We present the detailed case results of BIDeN on Task I. We show the results of $N = 2, 3, 4, 5, 6$ (BIDeN (2), BIDeN (3), BIDeN (4), BIDeN (5), BIDeN (6)) in Table~\ref{tab:2domain}, Table~\ref{tab:3domain}, Table~\ref{tab:4domain}, Table~\ref{tab:5domain}, and Table~\ref{tab:6domain}. These results are the extensions of Table 1 in the main paper. Note that due to the difference in precision, the PSNR results reported here are slightly different (less than $0.5\%$ difference) from the PSNR results reported in the main paper.

\begin{figure}[!htb]
     \centering
     \includegraphics[width = 5cm]
     {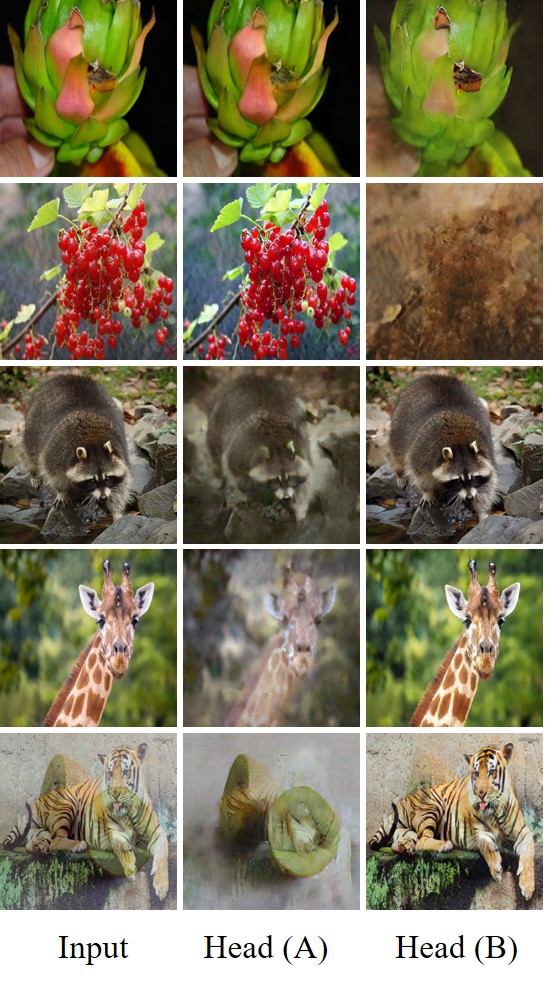}
     \caption{Qualitative results of BIDeN (2). Fruit (A), Animal (B). Row 1-2: a. Row 3-4: b. Row 5: ab}
     \label{fig:t1res2}
\end{figure}

\begin{figure}[!htb]
     \centering
     \includegraphics[width = 6cm]
     {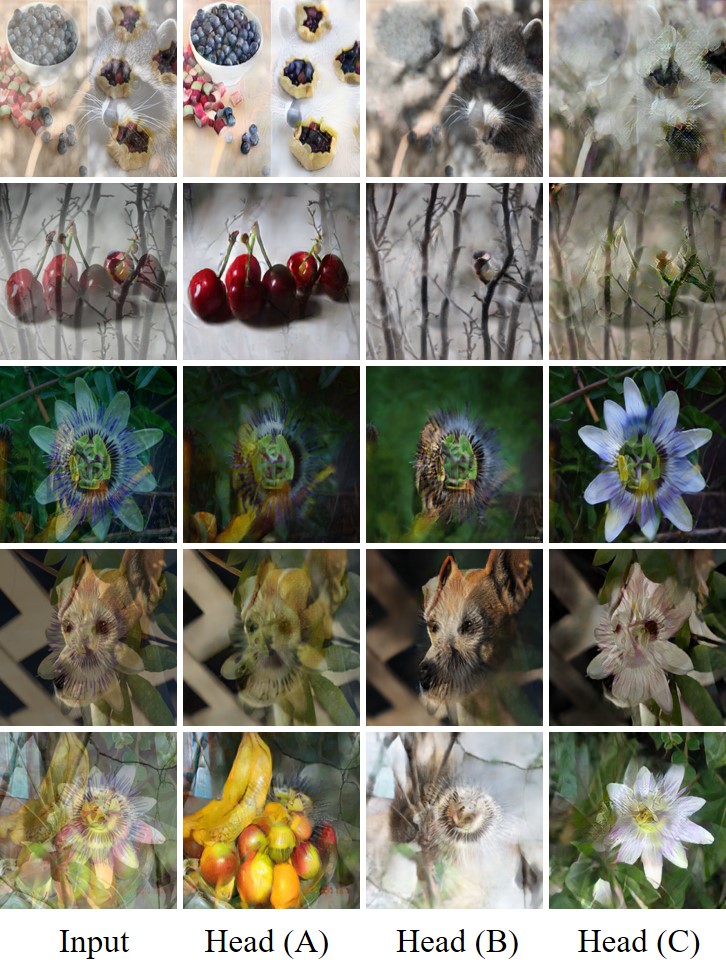}
     \caption{Qualitative results of BIDeN (3). Fruit (A), Animal (B), Flower (C). Row 1-2: ab. Row 3-4: bc. Row 5: abc}
     \label{fig:t1res3}
\end{figure}

\begin{figure}[!htb]
     \centering
     \includegraphics[width = 7cm]
     {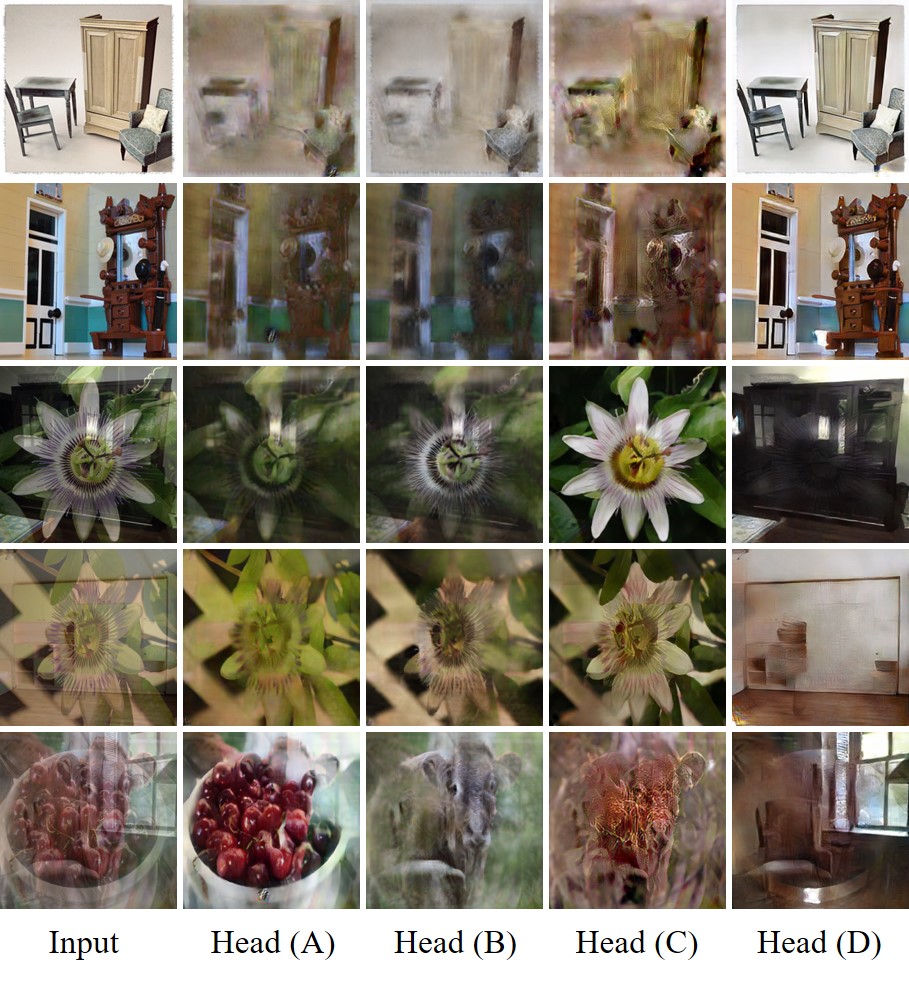}
     \caption{Qualitative results of BIDeN (4). Fruit (A), Animal (B), Flower (C), Furniture (D). Row 1-2: d. Row 3-4: cd. Row 5: abd}
     \label{fig:t1res4}
\end{figure}

\begin{figure}[!htb]
     \centering
     \includegraphics[width = 8cm]
     {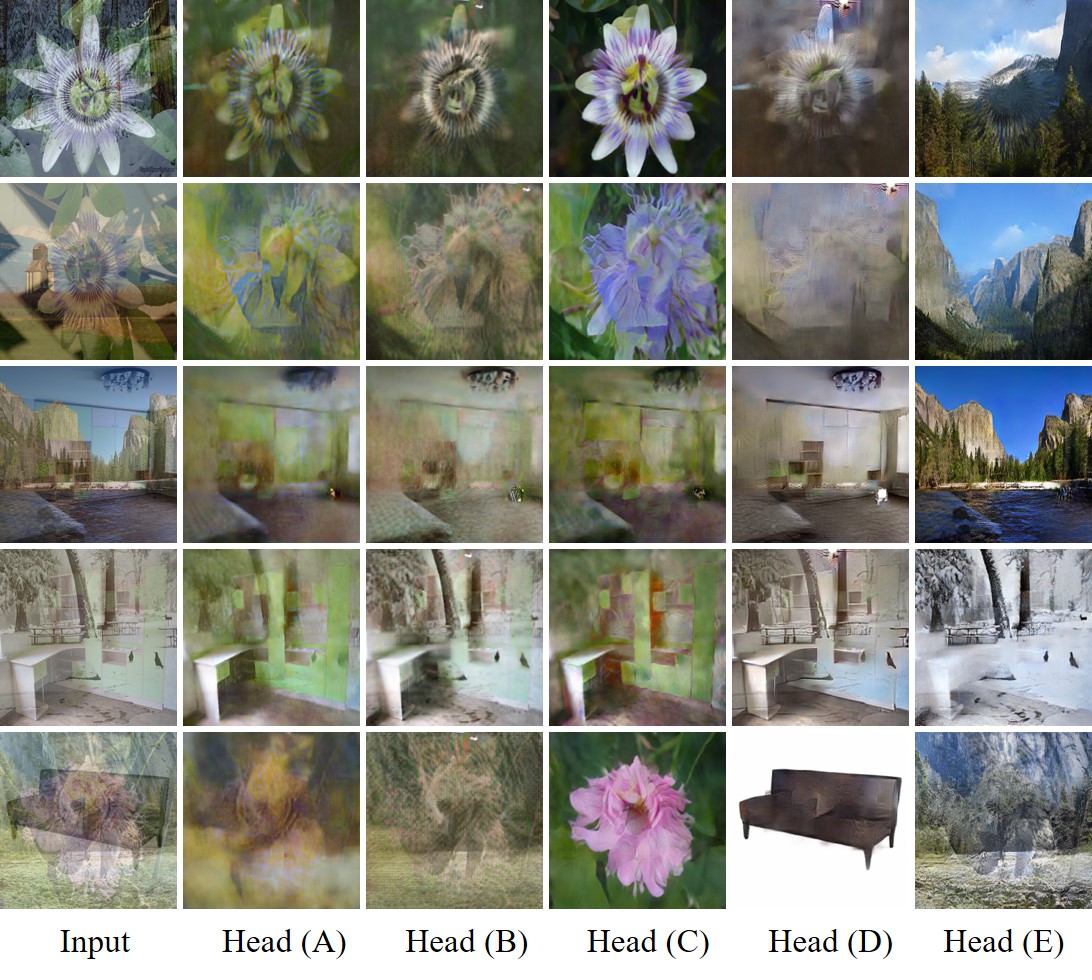}
     \caption{Qualitative results of BIDeN (5). Fruit (A), Animal (B), Flower (C), Furniture (D), Yosemite (E). Row 1-2: ce. Row 3-4: de. Row 5: bcde}
     \label{fig:t1res5}
\end{figure}

\begin{figure}[!htb]
     \centering
     \includegraphics[width = 8.3cm]
     {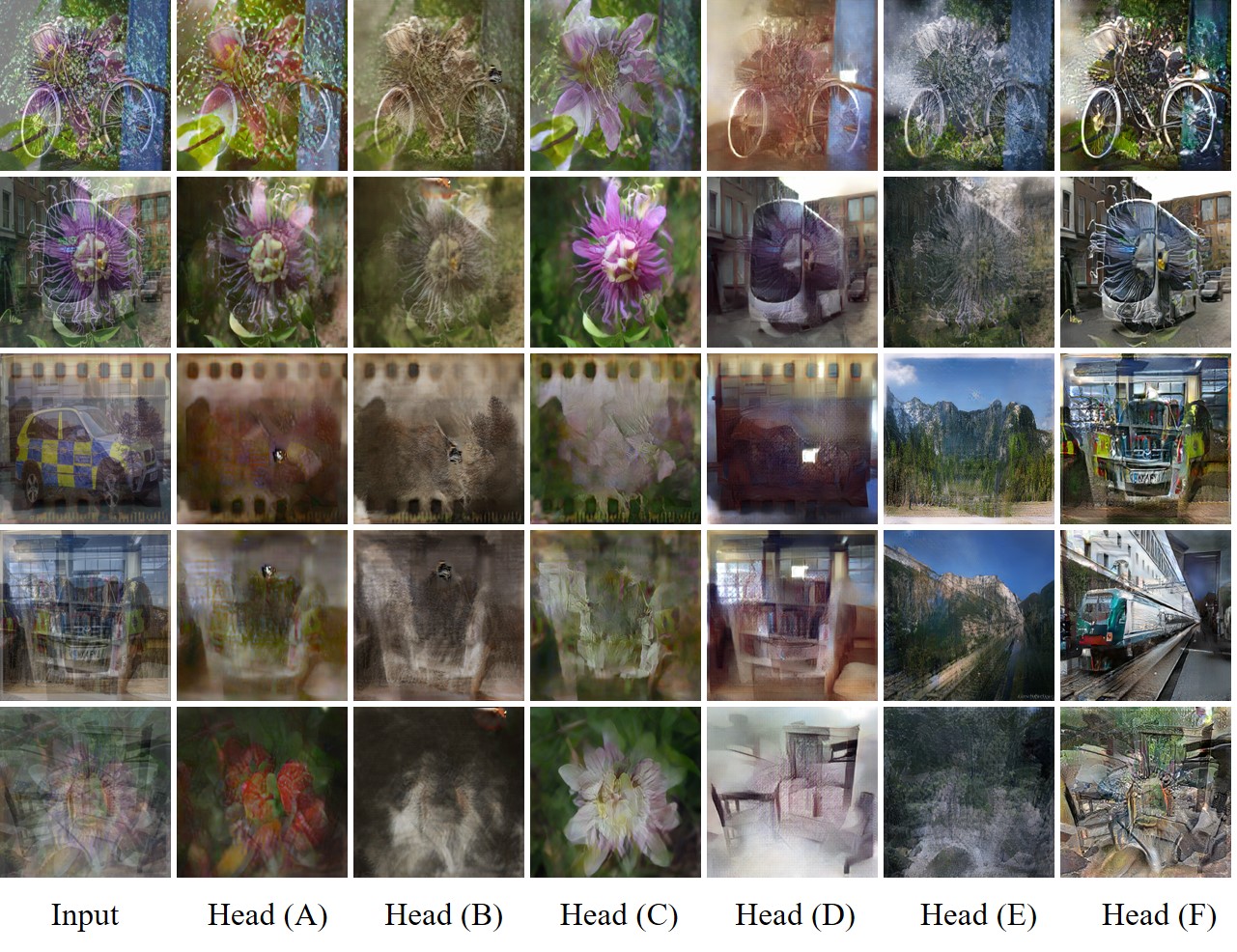}
     \caption{Qualitative results of BIDeN (6). Fruit (A), Animal (B), Flower (C), Furniture (D), Yosemite (E), Vehicle (F). Row 1-2: cf. Row 3-4: def. Row 5: abcdef}
     \label{fig:t1res6}
\end{figure}

\noindent \textbf{Qualitative results of BIDeN.} Here we present more qualitative results of BIDeN. We show the results of $N = 2, 3, 4, 5, 6$ (BIDeN (2), BIDeN (3), BIDeN (4), BIDeN (5), BIDeN (6)) in Figure~\ref{fig:t1res2}, Figure~\ref{fig:t1res3}, Figure~\ref{fig:t1res4}, Figure~\ref{fig:t1res5}, and Figure~\ref{fig:t1res6}.
The number of selected source components $L$ and the index set $I$ are randomly chosen. The eight source components in Task I are Fruit (A), Animal (B), Flower (C), Furniture (D), Yosemite (E), Vehicle (F), Vegetable (G), and CityScape (H). 

\begin{table}[!htbp]
  \centering
  \fontsize{7}{3}\selectfont
     \caption{Detailed case results of BIDeN (2) on Task I (Mixed image decomposition across multiple domains)}
    \begin{tabular}{lc|c|c}
    \toprule
    \textbf{Input}&\textbf{A}&\textbf{B}&\textbf{Acc}   \cr
    \midrule
    a& 25.26 & - &   0.940 \cr
    b& - & 25.11 &   0.933 \cr
    \cmidrule(lr){1-4} 
    ab& 20.09 & 19.93 &   1.000 \cr
    \cmidrule(lr){1-4} 
    \end{tabular}
     \label{tab:2domain}
\end{table}
\begin{table}[!htbp]
  \centering
  \fontsize{7}{3}\selectfont
       \caption{Detailed case results of BIDeN (3) on Task I (Mixed image decomposition across multiple domains)}
    \begin{tabular}{lc|c|c|cc}
    \toprule
    \textbf{Input}&\textbf{A}&\textbf{B}&\textbf{C}&\textbf{Acc}   \cr
    \midrule
    a& 24.04 & - & - &  0.906 \cr
    b& - & 23.48 & - &  0.953 \cr
    c& - & - & 22.74 &  0.756 \cr
    \cmidrule(lr){1-5} 
    ab& 19.07 & 18.78 & - &  0.833 \cr
    ac& 18.96 & - & 18.27 &  0.573 \cr
    bc& - & 18.31 & 18.25 &  0.763 \cr
    Avg& 19.01 & 18.54 & 18.26 &  0.723 \cr    
    \cmidrule(lr){1-5} 
    abc& 16.66 & 15.89 & 16.27 & 0.983 \cr 
    \cmidrule(lr){1-5}     
    \end{tabular}
     \label{tab:3domain}
\end{table}
\begin{table}[!htbp]
  \centering
  \fontsize{7}{3}\selectfont
       \caption{Detailed case results of BIDeN (4) on Task I (Mixed image decomposition across multiple domains)}
    \begin{tabular}{lc|c|c|c|cc}
    \toprule
    \textbf{Input}&\textbf{A}&\textbf{B}&\textbf{C}&\textbf{D}&\textbf{Acc}   \cr
    \midrule
    a& 23.07 & - & - & - & 0.896 \cr
    b& - & 22.61 & - & - & 0.886 \cr
    c& - & - & 21.93 & - &  0.770 \cr
    d& - & - & - & 21.88 &  0.933 \cr
    \cmidrule(lr){1-6} 
    ab& 18.22 & 18.06 & - & - &  0.710 \cr
    ac& 18.20 & - & 17.41 & - &  0.560 \cr
    ad& 18.85 & - & - & 18.51 & 0.783 \cr
    bc& - & 16.98 & 17.56 & - & 0.710 \cr    
    bc& - & 18.06 & - & 17.79 & 0.856 \cr    
    cd& - & - & 18.44 & 18.77 &  0.656 \cr
    Avg& 18.42 & 17.7 & 17.80 & 18.35 &  0.712 \cr   
    \cmidrule(lr){1-6} 
    abc& 16.12 & 15.46 & 15.70 & - &  0.660 \cr 
    abd& 16.47 & 15.95 & - & 16.40 &  0.676 \cr 
    acd& 16.36 & - & 16.11 & 16.78 & 0.396 \cr 
    bcd& - & 15.63 & 16.52 & 16.52 & 0.563 \cr
    Avg& 18.42 & 17.7 & 17.80 & 18.35 &  0.712 \cr   
    \cmidrule(lr){1-6}     
    abcd& 15.11 & 14.37 & 14.98 & 15.43 &  0.943 \cr 
    \cmidrule(lr){1-6}         
    \end{tabular}
     \label{tab:4domain}
\end{table}
\begin{table}[!htbp]
  \centering
  \fontsize{7}{3}\selectfont
       \caption{Detailed case results of BIDeN (5) on Task I (Mixed image decomposition across multiple domains)}
    \begin{tabular}{lc|c|c|c|c|cc}
    \toprule
    \textbf{Input}&\textbf{A}&\textbf{B}&\textbf{C}&\textbf{D}&\textbf{E}&\textbf{Acc}   \cr
    \midrule
    a& 22.75 & - & - & - & - & 0.840 \cr
    b& - & 22.02 & - & - & - & 0.853 \cr
    c& - & - & 22.30 & - & - & 0.553 \cr
    d& - & - & - & 22.52 & - & 0.930 \cr
    e& - & - & - & - & 20.77 & 0.923 \cr
    \cmidrule(lr){1-7} 
    ab& 17.68 & 17.30 & - & - & - & 0.676 \cr
    ac& 17.60 & - & 17.19 & - & - & 0.373 \cr
    ad& 18.34 & - & - & 18.26 & - & 0.720 \cr
    ae& 18.94 & - & - & - & 18.36 & 0.700 \cr
    bc& - & 16.98 & 17.56 & - & - & 0.633 \cr    
    bd& - & 17.21 & - & 17.60 & - & 0.716 \cr    
    be& - & 17.37 & - & - & 17.54 & 0.710 \cr 
    cd& - & - & 18.42 & 18.72 & - & 0.443 \cr   
    ce& - & - & 18.79 & - & 18.34 & 0.556 \cr    
    de& - & - & - & 18.18 & 17.94 & 0.796 \cr
    Avg& 18.14 & 17.21 & 17.99 & 18.19 & 18.04 & 0.632 \cr  
    \cmidrule(lr){1-7} 
    abc& 15.72 & 14.92 & 15.62 & - & - & 0.570 \cr 
    abd& 16.11 & 15.29 & - & 16.15 & - & 0.573 \cr 
    abe& 16.38 & 15.12 & - & - & 16.32 & 0.526 \cr     
    acd& 15.91 & - & 15.94 & 16.53 & - & 0.320 \cr        
    ace& 16.27 & - & 16.07 & - & 16.80 & 0.266 \cr   
    ade& 16.72 & - & - & 16.25 & 16.62 & 0.486 \cr       
    bcd& - & 15.04 & 16.46 & 16.32 & - & 0.576 \cr    
    bce& - & 14.94 & 16.62 & - & 16.38 & 0.553 \cr       
    bde& - & 15.24 & - & 16.07 & 16.15 & 0.736 \cr       
    cde& - & - & 17.08 & 16.48 & 16.52 & 0.366 \cr  
    Avg& 16.18 & 15.09 & 16.29 & 16.30 & 16.46 & 0.497 \cr 
    \cmidrule(lr){1-7}     
    abcd& 14.82 & 13.91 & 14.94 & 15.11 & - & 0.633 \cr 
    abce& 14.97 & 13.80 & 15.06 & - & 15.50 & 0.480 \cr   
    abde& 15.20 & 14.14 & - & 14.99 & 15.32 & 0.586 \cr 
    acde& 15.03 & - & 15.23 & 15.11 & 15.62 & 0.180 \cr   
    bcde& - & 13.92 & 15.74 & 15.06 & 15.34 & 0.570 \cr   
    Avg& 14.99 & 13.94 & 15.24 & 15.06 & 15.44 & 0.489 \cr 
    \cmidrule(lr){1-7}         
    abcde& 14.23 & 13.28 & 14.52 & 14.20 & 14.74 & 0.860 \cr     
    \cmidrule(lr){1-7}   
    \end{tabular}
     \label{tab:5domain}
\end{table}
\begin{table}[!htbp]
  \centering
  \fontsize{7}{3}\selectfont
       \caption{Detailed case results of BIDeN (6) on Task I (Mixed image decomposition across multiple domains)}
    \begin{tabular}{lc|c|c|c|c|c|cc}
    \toprule
    \textbf{Input}&\textbf{A}&\textbf{B}&\textbf{C}&\textbf{D}&\textbf{E}&\textbf{F}&\textbf{Acc}   \cr
    \midrule
    a& 22.82 & - & - & - & - & - & 0.850 \cr
    b& - & 21.80 & - & - & - & -& 0.826 \cr
    c& - & - & 22.61 & - & - & -& 0.823 \cr
    d& - & - & - & 22.50 & - & -& 0.890 \cr
    e& - & - & - & - & 21.09 & -& 0.910 \cr
    f& - & - & - & - & - & 21.97& 0.893 \cr
    \cmidrule(lr){1-8} 
    ab& 17.31 & 16.79 & - & - & - & -& 0.646 \cr
    ac& 17.10 & - & 16.84 & - & - & -& 0.526 \cr
    ad& 18.14 & - & - & 17.66 & - & -& 0.613 \cr
    ae& 18.59 & - & - & - & 18.24 & -& 0.676 \cr
    af& 18.24 & - & - & - & - & 17.76& 0.583 \cr
    bc& - & 16.60 & 17.46 & - & - & -& 0.613 \cr    
    bd& - & 16.95 & - & 17.05 & - & -& 0.593 \cr    
    be& - & 16.81 & - & - & 17.30 & -& 0.660 \cr 
    bf& - & 16.90 & - & - & - & 17.04& 0.510 \cr 
    cd& - & - & 18.23 & 18.11 & - & -& 0.480 \cr   
    ce& - & - & 18.53 & - & 18.16 & -& 0.670 \cr    
    cf& - & - & 18.13 & - & - & 17.48& 0.583 \cr    
    de& - & - & - & 17.43 & 17.86 & -& 0.653 \cr 
    df& - & - & - & 16.18 & - & 16.01& 0.686 \cr     
    ef& - & - & - & - & 17.42 & 16.74& 0.673 \cr  
    Avg& 17.87 & 16.81 & 17.83 & 17.28 & 17.79 & 17.00& 0.654 \cr    
    \cmidrule(lr){1-8} 
    abc& 15.29 & 14.56 & 15.34 & - & - & -& 0.606 \cr 
    abd& 15.83 & 14.95 & - & 15.51 & - & -& 0.390 \cr 
    abe& 16.10 & 14.59 & - & - & 16.04 & -& 0.513 \cr
    abf& 15.89 & 14.78 & - & - & - & 15.61& 0.333 \cr     
    acd& 15.49 & - & 15.56 & 15.98 & - & -& 0.313 \cr        
    ace& 15.88 & - & 15.67 & - & 16.55 & -& 0.350 \cr   
    acf& 15.64 & - & 15.46 & - & - & 15.61& 0.303 \cr   
    ade& 16.46 & - & - & 15.59 & 16.45 & -& 0.330 \cr 
    adf& 16.15 & - & - & 14.60 & - & 14.73& 0.450 \cr
    aef& 16.44 & - & - & - & 15.99 & 15.18& 0.373 \cr  
    bce& - & 14.68 & 16.24 & 15.71 & - & -& 0.370 \cr    
    bce& - & 14.43 & 16.43 & - & 16.08 & -& 0.500 \cr   
    bcf& - & 14.69 & 16.04 & - & - & 15.42& 0.346 \cr      
    bde& - & 14.90 & - & 15.45 & 15.89 & -& 0.516 \cr     
    bdf& - & 14.87 & - & 14.60 & - & 14.52& 0.456 \cr
    bef& - & 14.81 & - & - & 15.60 & 15.04& 0.400 \cr    
    cde& - & - & 16.82 & 15.84 & 16.30 & -& 0.253 \cr   
    cdf& - & - & 16.48 & 14.86 & - & 14.59& 0.383 \cr  
    cef& - & - & 16.64 & - & 15.91 & 15.15& 0.443 \cr 
    def& - & - & - & 14.70 & 15.80 & 14.26& 0.506 \cr
    Avg& 15.92 & 14.72 & 16.06 & 15.28 & 16.06& 16.71& 0.439 \cr    
    \cmidrule(lr){1-8}     
    abcd& 14.45 & 13.69 & 14.61 & 14.62 & - & -& 0.420 \cr 
    abce& 14.63 & 13.51 & 14.76 & - & 15.20 & -& 0.540 \cr   
    abcf& 14.57 & 13.65 & 14.54 & - & - & 14.34& 0.313 \cr   
    abde& 15.02 & 13.79 & - & 14.36 & 15.05 & -& 0.323 \cr 
    abdf& 14.90 & 13.80 & - & 13.77 & - & 13.81& 0.373 \cr 
    abef& 15.13 & 13.70 & - & - & 14.77 & 14.17& 0.276 \cr     
    acde& 14.71 & - & 14.88 & 14.60 & 15.34 & -& 0.126 \cr 
    acdf& 14.61 & - & 14.67 & 13.97 & -& 13.75& 0.220 \cr  
    acef& 14.80 & - & 14.81 & - & 14.99& 14.14& 0.193 \cr 
    adef& 15.13 & - & - & 14.81 & 15.00& 13.50& 0.293 \cr     
    bcde& - & 13.64 & 15.55 & 14.56 & 14.98 & -& 0.296 \cr   
    bcdf& - & 13.76 & 15.26 & 13.96 &- & 13.66& 0.336 \cr       
    bcef& - & 13.63 & 15.35 & - &14.84 & 14.10& 0.356 \cr       
    bdef& - & 13.86 & - & 13.78 &14.74 & 13.49& 0.543 \cr       
    cdef& - & - & 15.62 & 13.99 &14.99 & 13.44& 0.206 \cr
    Avg& 14.79 & 13.70 & 15.00 & 14.24 & 14.99& 13.84& 0.321 \cr   
    \cmidrule(lr){1-8}         
    abcde& 13.97 & 13.12 & 14.24 & 13.73 & 14.41 & -& 0.350 \cr 
    abcdf& 13.93 & 13.22 & 14.05 & 13.38 & - & 13.14& 0.543 \cr     
    abcef& 14.04 & 13.09 & 14.17 & - & 14.23 & 13.41& 0.346 \cr       
    abedf& 14.30 & 13.25 & - & 13.24 & 14.20 & 12.99& 0.516 \cr   
    acdef& 14.03 & - & 14.28 & 13.40 & 14.37 & 12.94& 0.140 \cr      
    bcdef& - & 13.18 & 14.76 & 13.40 & 14.19 & 12.92& 0.356 \cr 
    Avg& 14.04 & 13.17 & 14.30 & 13.43 & 14.28& 13.08& 0.375 \cr  
    \cmidrule(lr){1-8}   
    abcdef& 13.54 & 12.86 & 13.82 & 12.95 & 13.80 & 12.57& 0.846 \cr   
    \cmidrule(lr){1-8}   
    \end{tabular}
     \label{tab:6domain}
\end{table}

\clearpage

\subsection{Additional results of Task II}
\begin{figure*}[!htb]
     \centering
     \includegraphics[width = 12cm]
     {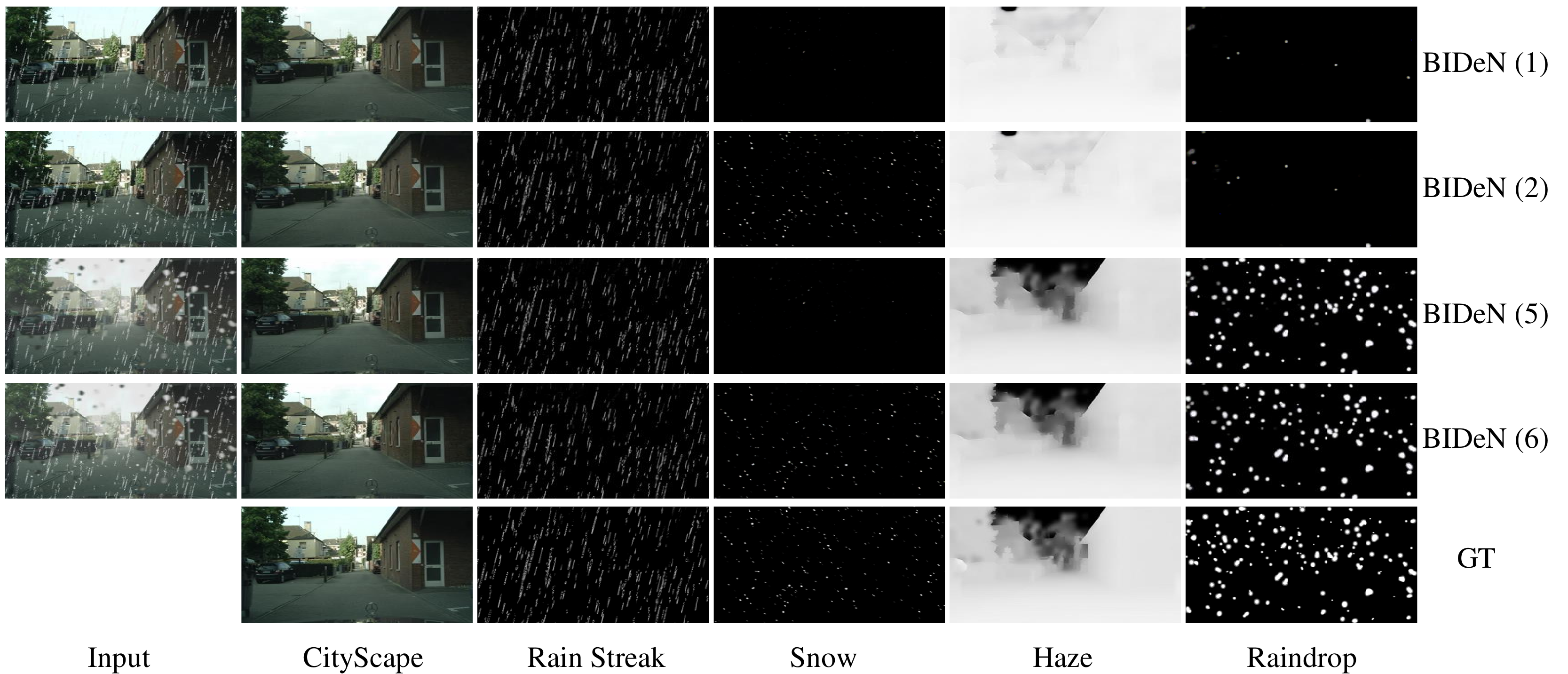}
     \caption{CityScape, masks (Rain Streak, Snow, Raindrop), and transmission map (Haze) generated by BIDeN for case (1), case (2), case (5), and case (6). Case (1): rain streak, case (2): rain streak + snow, case (5): rain streak + moderate haze + raindrop, case (6): rain streak + snow + moderate haze + raindrop }
     \label{fig:supprain}
\end{figure*}

\begin{figure*}[!htbp]
  \begin{minipage}[t]{0.24\linewidth} 
    \centering 
    \text{Input}
  \end{minipage} 
    \begin{minipage}[t]{0.24\linewidth} 
    \centering 
    \text{MPRNet}
  \end{minipage} 
    \begin{minipage}[t]{0.24\linewidth} 
    \centering 
      \text{ BIDeN}
  \end{minipage} 
    \begin{minipage}[t]{0.24\linewidth} 
    \centering 
        \text{ GT}
  \end{minipage}
  \\
  \begin{minipage}[t]{0.24\linewidth} 
    \centering 
    \includegraphics[width=1.1in, height=0.55in]{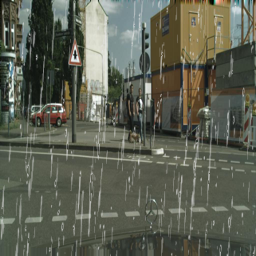}
  \end{minipage} 
    \begin{minipage}[t]{0.24\linewidth} 
    \centering 
        \includegraphics[width=1.1in, height=0.55in]{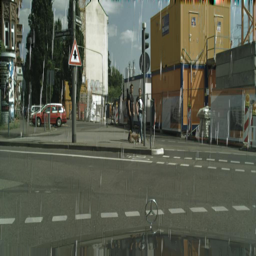}
  \end{minipage}
   \begin{minipage}[t]{0.24\linewidth} 
    \centering 
    \includegraphics[width=1.1in, height=0.55in]{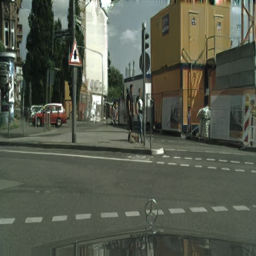}
  \end{minipage} 
    \begin{minipage}[t]{0.24\linewidth} 
    \centering 
        \includegraphics[width=1.1in, height=0.55in]{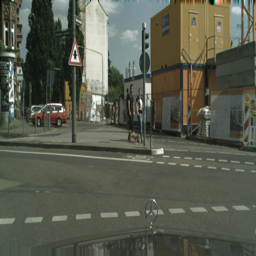}
  \end{minipage} 
\\
  \begin{minipage}[t]{0.24\linewidth} 
    \centering 
    \includegraphics[width=1.1in, height=0.55in]{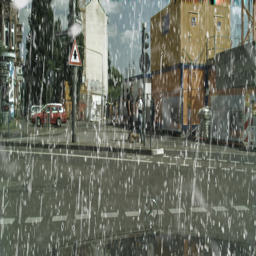}
  \end{minipage} 
    \begin{minipage}[t]{0.24\linewidth} 
    \centering 
        \includegraphics[width=1.1in, height=0.55in]{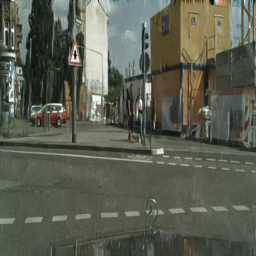}
  \end{minipage}
   \begin{minipage}[t]{0.24\linewidth} 
    \centering 
    \includegraphics[width=1.1in, height=0.55in]{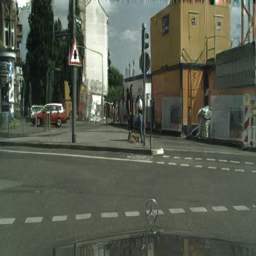}
  \end{minipage} 
    \begin{minipage}[t]{0.24\linewidth} 
    \centering 
        \includegraphics[width=1.1in, height=0.55in]{figures/rain3/G1.png}
  \end{minipage} 
\\
  \begin{minipage}[t]{0.24\linewidth} 
    \centering 
    \includegraphics[width=1.1in, height=0.55in]{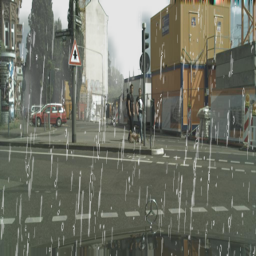}
  \end{minipage} 
    \begin{minipage}[t]{0.24\linewidth} 
    \centering 
        \includegraphics[width=1.1in, height=0.55in]{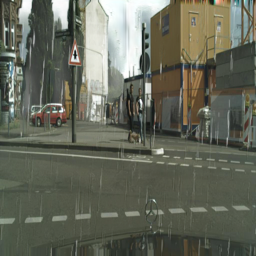}
  \end{minipage}
   \begin{minipage}[t]{0.24\linewidth} 
    \centering 
    \includegraphics[width=1.1in, height=0.55in]{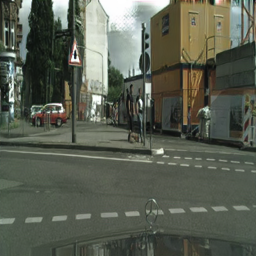}
  \end{minipage} 
    \begin{minipage}[t]{0.24\linewidth} 
    \centering 
        \includegraphics[width=1.1in, height=0.55in]{figures/rain3/G1.png}
  \end{minipage} 
\\
  \begin{minipage}[t]{0.24\linewidth} 
    \centering 
    \includegraphics[width=1.1in, height=0.55in]{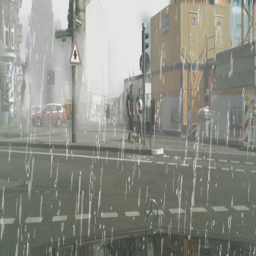}
  \end{minipage} 
    \begin{minipage}[t]{0.24\linewidth} 
    \centering 
        \includegraphics[width=1.1in, height=0.55in]{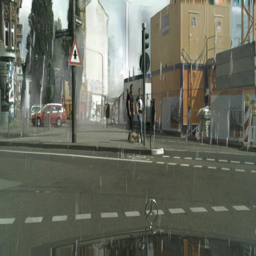}
  \end{minipage}
   \begin{minipage}[t]{0.24\linewidth} 
    \centering 
    \includegraphics[width=1.1in, height=0.55in]{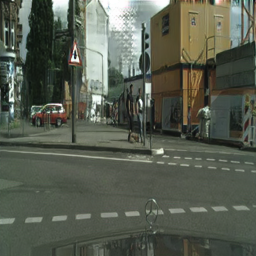}
  \end{minipage} 
    \begin{minipage}[t]{0.24\linewidth} 
    \centering 
        \includegraphics[width=1.1in, height=0.55in]{figures/rain3/G1.png}
  \end{minipage} 
\\
  \begin{minipage}[t]{0.24\linewidth} 
    \centering 
    \includegraphics[width=1.1in, height=0.55in]{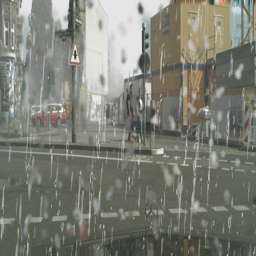}
  \end{minipage} 
    \begin{minipage}[t]{0.24\linewidth} 
    \centering 
        \includegraphics[width=1.1in, height=0.55in]{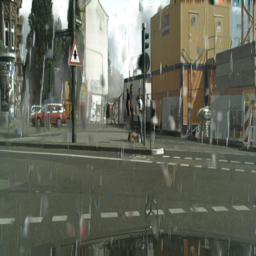}
  \end{minipage}
   \begin{minipage}[t]{0.24\linewidth} 
    \centering 
    \includegraphics[width=1.1in, height=0.55in]{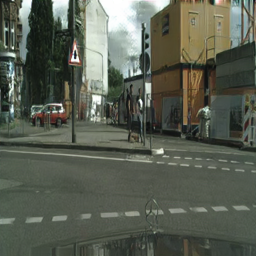}
  \end{minipage} 
    \begin{minipage}[t]{0.24\linewidth} 
    \centering 
        \includegraphics[width=1.1in, height=0.55in]{figures/rain3/G1.png}
  \end{minipage} 
\\
  \begin{minipage}[t]{0.24\linewidth} 
    \centering 
    \includegraphics[width=1.1in, height=0.55in]{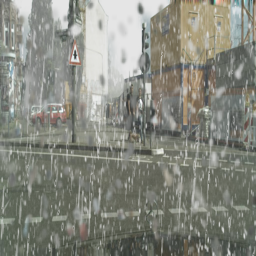}
  \end{minipage} 
    \begin{minipage}[t]{0.24\linewidth} 
    \centering 
        \includegraphics[width=1.1in, height=0.55in]{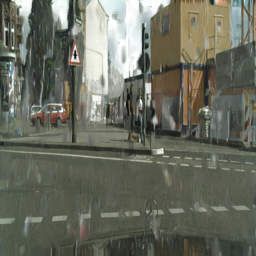}
  \end{minipage}
   \begin{minipage}[t]{0.24\linewidth} 
    \centering 
    \includegraphics[width=1.1in, height=0.55in]{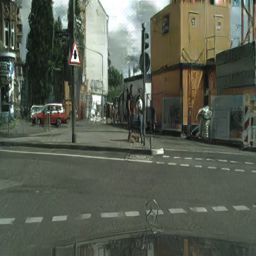}
  \end{minipage} 
    \begin{minipage}[t]{0.24\linewidth} 
    \centering 
        \includegraphics[width=1.1in, height=0.55in]{figures/rain3/G1.png}
  \end{minipage} 
  \caption{Additional results of Task II.A (Real-scenario deraining in driving). Row 1-6 presents 6 cases of a same scene. The 6 cases are (1): rain streak, (2): rain streak + snow, (3): rain streak + light haze, (4): rain streak + heavy haze, (5): rain streak + moderate haze + raindrop, (6) rain streak + snow + moderate haze + raindrop. BIDeN remove all components of rain efficiently while MPRNet leaves some components that are not completely removed}
  \label{fig:rain3}
\end{figure*}

\begin{figure*}[!htbp]
  \begin{minipage}[t]{0.24\linewidth} 
    \centering 
    \text{Input}
  \end{minipage} 
    \begin{minipage}[t]{0.24\linewidth} 
    \centering 
    \text{MPRNet}
  \end{minipage} 
    \begin{minipage}[t]{0.24\linewidth} 
    \centering 
      \text{ BIDeN}
  \end{minipage} 
    \begin{minipage}[t]{0.24\linewidth} 
    \centering 
        \text{ GT}
  \end{minipage}
  \\
  \begin{minipage}[t]{0.24\linewidth} 
    \centering 
    \includegraphics[width=1.1in, height=0.55in]{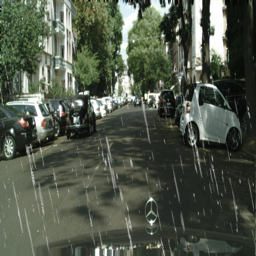}
  \end{minipage} 
    \begin{minipage}[t]{0.24\linewidth} 
    \centering 
        \includegraphics[width=1.1in, height=0.55in]{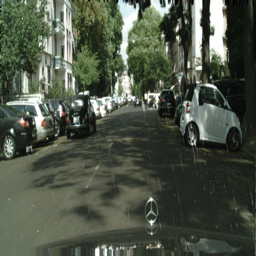}
  \end{minipage}
   \begin{minipage}[t]{0.24\linewidth} 
    \centering 
    \includegraphics[width=1.1in, height=0.55in]{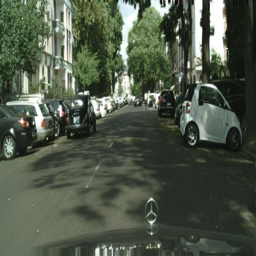}
  \end{minipage} 
    \begin{minipage}[t]{0.24\linewidth} 
    \centering 
        \includegraphics[width=1.1in, height=0.55in]{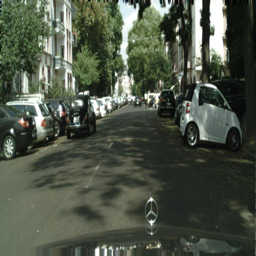}
  \end{minipage} 
\\
  \begin{minipage}[t]{0.24\linewidth} 
    \centering 
    \includegraphics[width=1.1in, height=0.55in]{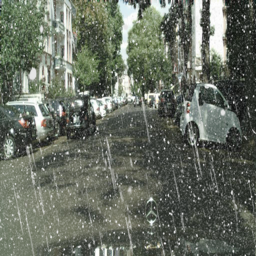}
  \end{minipage} 
    \begin{minipage}[t]{0.24\linewidth} 
    \centering 
        \includegraphics[width=1.1in, height=0.55in]{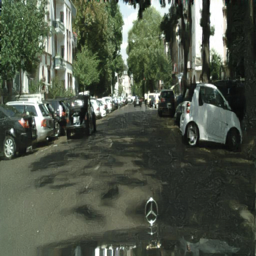}
  \end{minipage}
   \begin{minipage}[t]{0.24\linewidth} 
    \centering 
    \includegraphics[width=1.1in, height=0.55in]{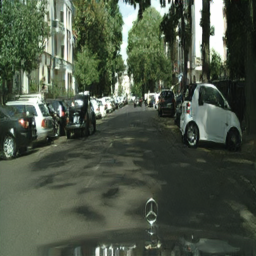}
  \end{minipage} 
    \begin{minipage}[t]{0.24\linewidth} 
    \centering 
        \includegraphics[width=1.1in, height=0.55in]{figures/rain3/G2.png}
  \end{minipage} 
\\
  \begin{minipage}[t]{0.24\linewidth} 
    \centering 
    \includegraphics[width=1.1in, height=0.55in]{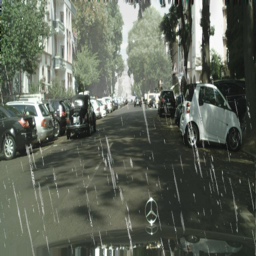}
  \end{minipage} 
    \begin{minipage}[t]{0.24\linewidth} 
    \centering 
        \includegraphics[width=1.1in, height=0.55in]{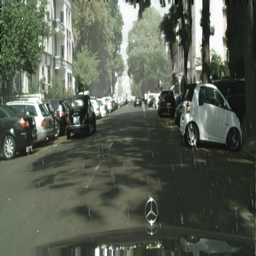}
  \end{minipage}
   \begin{minipage}[t]{0.24\linewidth} 
    \centering 
    \includegraphics[width=1.1in, height=0.55in]{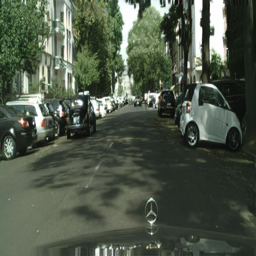}
  \end{minipage} 
    \begin{minipage}[t]{0.24\linewidth} 
    \centering 
        \includegraphics[width=1.1in, height=0.55in]{figures/rain3/G2.png}
  \end{minipage} 
\\
  \begin{minipage}[t]{0.24\linewidth} 
    \centering 
    \includegraphics[width=1.1in, height=0.55in]{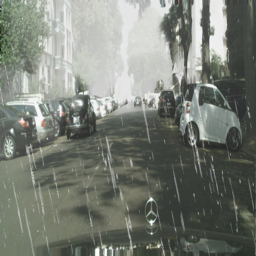}
  \end{minipage} 
    \begin{minipage}[t]{0.24\linewidth} 
    \centering 
        \includegraphics[width=1.1in, height=0.55in]{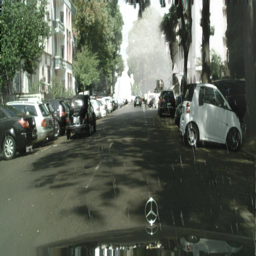}
  \end{minipage}
   \begin{minipage}[t]{0.24\linewidth} 
    \centering 
    \includegraphics[width=1.1in, height=0.55in]{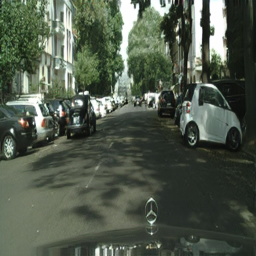}
  \end{minipage} 
    \begin{minipage}[t]{0.24\linewidth} 
    \centering 
        \includegraphics[width=1.1in, height=0.55in]{figures/rain3/G2.png}
  \end{minipage} 
\\
  \begin{minipage}[t]{0.24\linewidth} 
    \centering 
    \includegraphics[width=1.1in, height=0.55in]{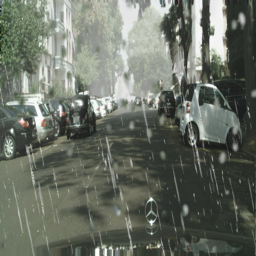}
  \end{minipage} 
    \begin{minipage}[t]{0.24\linewidth} 
    \centering 
        \includegraphics[width=1.1in, height=0.55in]{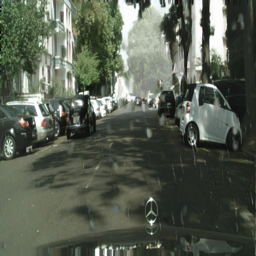}
  \end{minipage}
   \begin{minipage}[t]{0.24\linewidth} 
    \centering 
    \includegraphics[width=1.1in, height=0.55in]{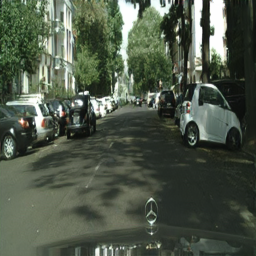}
  \end{minipage} 
    \begin{minipage}[t]{0.24\linewidth} 
    \centering 
        \includegraphics[width=1.1in, height=0.55in]{figures/rain3/G2.png}
  \end{minipage} 
\\
  \begin{minipage}[t]{0.24\linewidth} 
    \centering 
    \includegraphics[width=1.1in, height=0.55in]{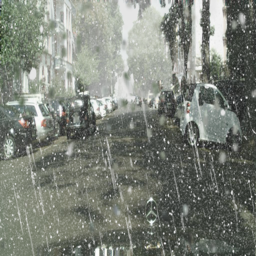}
  \end{minipage} 
    \begin{minipage}[t]{0.24\linewidth} 
    \centering 
        \includegraphics[width=1.1in, height=0.55in]{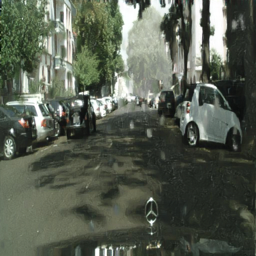}
  \end{minipage}
   \begin{minipage}[t]{0.24\linewidth} 
    \centering 
    \includegraphics[width=1.1in, height=0.55in]{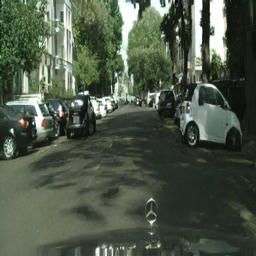}
  \end{minipage} 
    \begin{minipage}[t]{0.24\linewidth} 
    \centering 
        \includegraphics[width=1.1in, height=0.55in]{figures/rain3/G2.png}
  \end{minipage} 
  \caption{Additional results of Task II.A (Real-scenario deraining in driving). Row 1-6 presents 6 cases of a same scene. The 6 cases are (1): rain streak, (2): rain streak + snow, (3): rain streak + light haze, (4): rain streak + heavy haze, (5): rain streak + moderate haze + raindrop, (6) rain streak + snow + moderate haze + raindrop. BIDeN remove all components of rain efficiently while MPRNet leaves some components that are not completely removed}
  \label{fig:rain4}
\end{figure*}

For Task II.A (Real-scenario deraining in driving), more qualitative results are provided. 
Visual examples of CityScapes/masks/transmission maps generated by BIDeN are shown in Figure~\ref{fig:supprain}. We present more qualitative comparisons between BIDeN and MPRNet~\cite{Zamir2021MPRNet} in Figure~\ref{fig:rain3} and Figure~\ref{fig:rain4}. The comparison presents the results of 6 cases from the same scene. 

For the default training setting on Task II.A, the probabilities of every component to be selected are 1.0, 0.5, 0.5, 0.5 for rain streak, snow, raindrop, and haze. Moreover, we train both BIDeN and MPRNet again setting the possibility of the rain streak component to be selected as 0.8. The quantitative results of BIDeN and the comparison between MPRNet are provided in Table~\ref{tab:s2} and Table~\ref{tab:s22}. Compared to BIDeN trained under the default training setting of Task II.A, BIDeN performs better when the possibility of the rain streak component to be selected is set to 0.8. 

\begin{table*}[!htbp]
  \centering
  \fontsize{7}{3}\selectfont
       \caption{Results of BIDeN on Task II.A (Real-scenario deraining in driving). The possibility of the rain streak component to be selected is 0.8. We employ PSNR and SSIM metrics for both CityScape images, masks, and transmission maps. We report the results for 6 test cases as presented in Figure 1 of the main paper, the 6 cases are (1): rain streak, (2): rain streak + snow, (3): rain streak + light haze, (4): rain streak + heavy haze, (5): rain streak + moderate haze + raindrop, (6) rain streak + snow + moderate haze + raindrop. Note that only haze is divided into light/moderate/heavy intensities. Both training set and test set of Rain Streak, Snow, and Raindrop already consist of different intensities}
    \begin{tabular}{lc|c|c|c|c|c|c|c|c|c|c}
    \toprule
     \Rows{Method}&\multicolumn{2}{c|}{CityScape}&\multicolumn{2}{c|}{Rain Streak}&\multicolumn{2}{c|}{Snow}&\multicolumn{2}{c|}{Haze}&\multicolumn{2}{c|}{Raindrop}& \Rows{Acc $\uparrow$}\cr
    &PSNR$\uparrow$ & SSIM$\uparrow$ & PSNR$\uparrow$ & SSIM$\uparrow$& PSNR$\uparrow$ & SSIM$\uparrow$ & PSNR$\uparrow$ & SSIM$\uparrow$ & PSNR$\uparrow$ & SSIM$\uparrow$\cr
    \midrule
    BIDeN (1)& 33.30&0.930 & 31.55&0.917 & - & - & - & - & - & - & 1.0 \cr
    BIDeN (2)& 29.55&0.896 & 28.80&0.836 & 25.74&0.689 & - & - & - & - & 0.992 \cr    
    BIDeN (3)& 29.38&0.919 & 30.98&0.907 & - & - & 31.11&0.956 & - & -& 0.994 \cr
    BIDeN (4)& 27.56&0.899 & 30.39&0.895 & - & - & 31.83&0.944 & - & -& 0.994 \cr   
    BIDeN (5)& 27.89&0.900 & 30.17&0.891 & - &- & 30.73&0.945 & 22.32&0.904 & 0.996 \cr
    BIDeN (6)& 27.05&0.869 & 28.05&0.815 & 24.81&0.653 & 30.02&0.940 & 21.55&0.888 & 0.990 \cr     
    \bottomrule
    \end{tabular}
     \label{tab:s2}
\end{table*}

\begin{table*}[!htbp]
  \centering
  \fontsize{7}{3}\selectfont
      \caption{Comparison on Task II.A (Real-scenario deraining in driving) between BIDeN and MPRNet~\cite{Zamir2021MPRNet}. The possibility of the rain streak component to be selected is 0.8. MPRNet shows superior results for case (1) and case (2). In contrast, BIDeN is better at other cases. For the details of 6 test cases, please refer to Table~\ref{tab:s2} and Figure 1 of the main paper}
    \begin{tabular}{lc|c|c|c|c|c}
    \toprule
     \Rows{Case}& \multicolumn{2}{c|}{Input} & \multicolumn{2}{c|}{MPRNet} & \multicolumn{2}{c}{BIDeN}   \cr
    &PSNR$\uparrow$ & SSIM$\uparrow$ & PSNR$\uparrow$ & SSIM$\uparrow$& PSNR$\uparrow$ & SSIM$\uparrow$\cr
    \midrule
    \enspace (1)& 25.69&0.786 & 33.03&0.941  & 33.30&0.930 \cr
    \enspace (2)& 18.64&0.564 & 30.44&0.902  & 29.55&0.896 \cr
    \enspace (3)& 17.45&0.712 & 23.95&0.897  & 29.38&0.919 \cr    
    \enspace (4)& 11.12&0.571 & 17.32&0.810  & 27.56&0.899 \cr  
    \enspace (5)& 14.05&0.616 & 20.75&0.839  & 27.89&0.900 \cr  
    \enspace (6)& 12.38&0.461 & 19.74&0.798  & 27.05&0.869 \cr        
    \bottomrule
    \end{tabular}
     \label{tab:s22}
\end{table*}
\clearpage

\subsection{Additional results of Task III}
\label{appendix: results}
\begin{figure}[!htbp]
  \begin{minipage}[t]{0.19\linewidth} 
    \centering 
    \text{\small Input}
  \end{minipage} 
    \begin{minipage}[t]{0.19\linewidth} 
    \centering 
    \text{\small DHAN}
  \end{minipage} 
    \begin{minipage}[t]{0.19\linewidth} 
    \centering 
      \text{\small Auto-Exp}
  \end{minipage} 
    \begin{minipage}[t]{0.19\linewidth} 
    \centering 
        \text{\small BIDeN}
  \end{minipage}
      \begin{minipage}[t]{0.19\linewidth} 
    \centering 
        \text{\small GT}
  \end{minipage}
  \\
  \begin{minipage}[t]{0.19\linewidth} 
    \centering 
    \includegraphics[width=0.62in, height=0.62in]{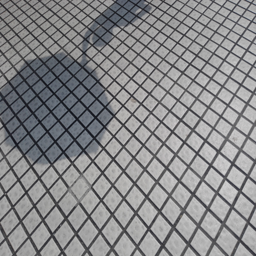}
  \end{minipage} 
    \begin{minipage}[t]{0.19\linewidth} 
    \centering 
        \includegraphics[width=0.62in, height=0.62in]{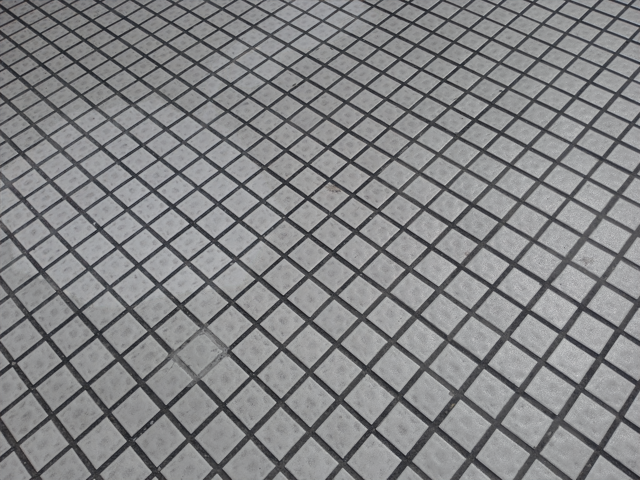}
  \end{minipage}
     \begin{minipage}[t]{0.19\linewidth} 
    \centering 
    \includegraphics[width=0.62in, height=0.62in]{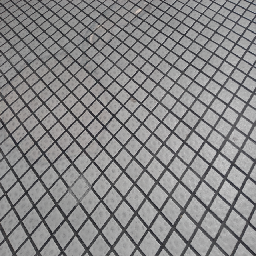}
  \end{minipage} 
   \begin{minipage}[t]{0.19\linewidth} 
    \centering 
    \includegraphics[width=0.62in, height=0.62in]{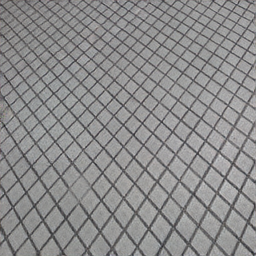}
  \end{minipage} 
    \begin{minipage}[t]{0.19\linewidth} 
    \centering 
        \includegraphics[width=0.62in, height=0.62in]{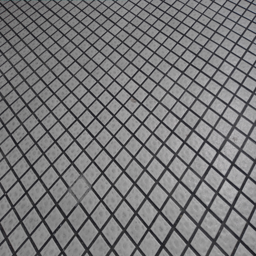}
  \end{minipage} 
  \\
    \begin{minipage}[t]{0.19\linewidth} 
    \centering 
    \includegraphics[width=0.62in, height=0.62in]{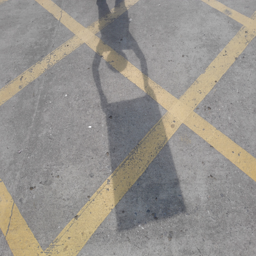}
  \end{minipage} 
    \begin{minipage}[t]{0.19\linewidth} 
    \centering 
        \includegraphics[width=0.62in, height=0.62in]{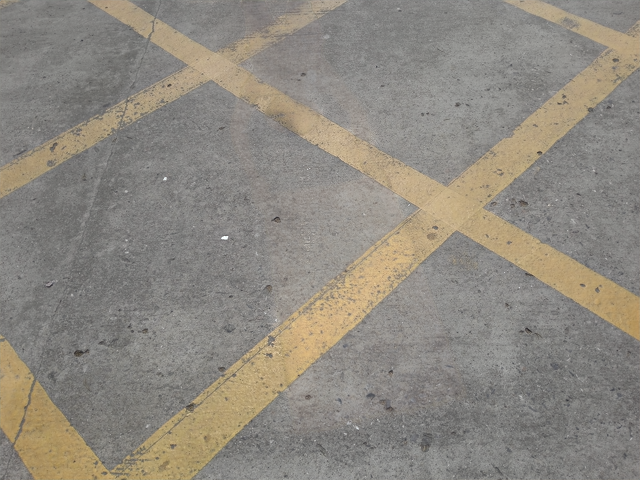}
  \end{minipage}
     \begin{minipage}[t]{0.19\linewidth} 
    \centering 
    \includegraphics[width=0.62in, height=0.62in]{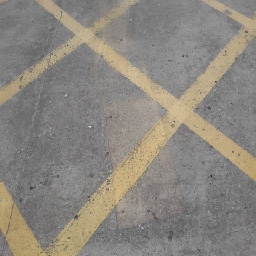}
  \end{minipage} 
   \begin{minipage}[t]{0.19\linewidth} 
    \centering 
    \includegraphics[width=0.62in, height=0.62in]{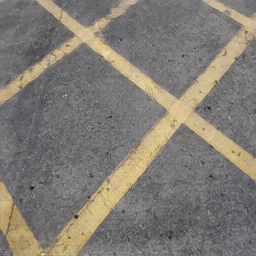}
  \end{minipage} 
    \begin{minipage}[t]{0.19\linewidth} 
    \centering 
        \includegraphics[width=0.62in, height=0.62in]{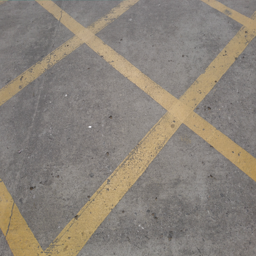}
  \end{minipage} 
  \\
      \begin{minipage}[t]{0.19\linewidth} 
    \centering 
    \includegraphics[width=0.62in, height=0.62in]{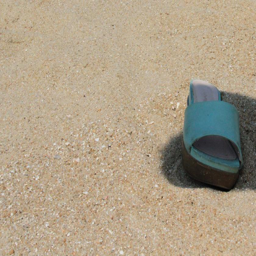}
  \end{minipage} 
    \begin{minipage}[t]{0.19\linewidth} 
    \centering 
        \includegraphics[width=0.62in, height=0.62in]{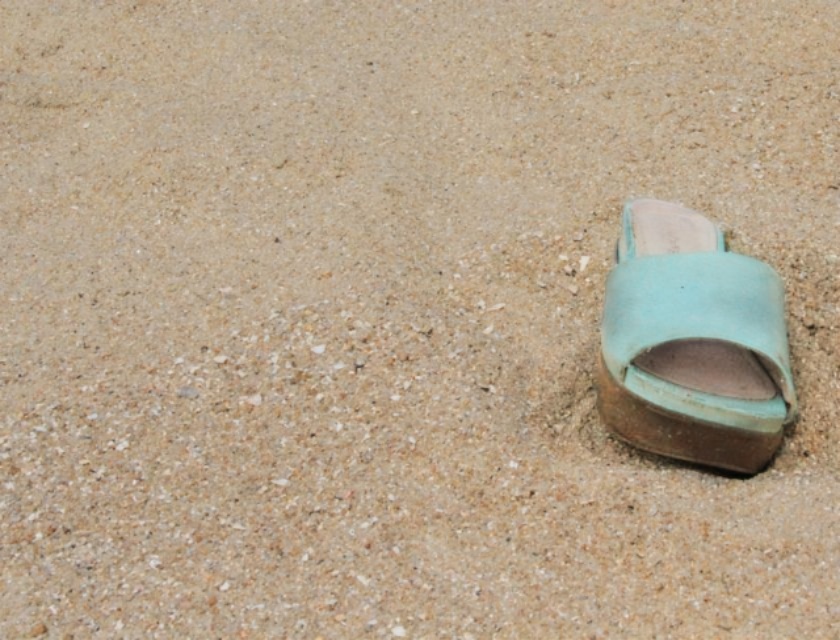}
  \end{minipage}
     \begin{minipage}[t]{0.19\linewidth} 
    \centering 
    \includegraphics[width=0.62in, height=0.62in]{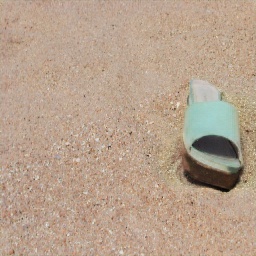}
  \end{minipage} 
   \begin{minipage}[t]{0.19\linewidth} 
    \centering 
    \includegraphics[width=0.62in, height=0.62in]{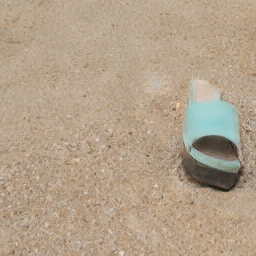}
  \end{minipage} 
    \begin{minipage}[t]{0.19\linewidth} 
    \centering 
        \includegraphics[width=0.62in, height=0.62in]{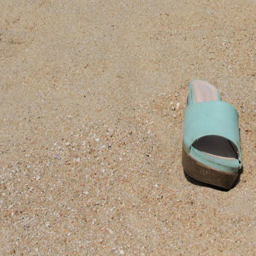}
  \end{minipage} 
  \\
      \begin{minipage}[t]{0.19\linewidth} 
    \centering 
    \includegraphics[width=0.62in, height=0.62in]{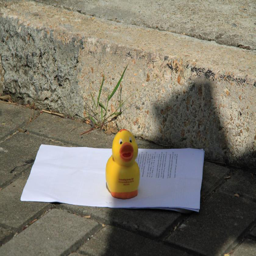}
  \end{minipage} 
    \begin{minipage}[t]{0.19\linewidth} 
    \centering 
        \includegraphics[width=0.62in, height=0.62in]{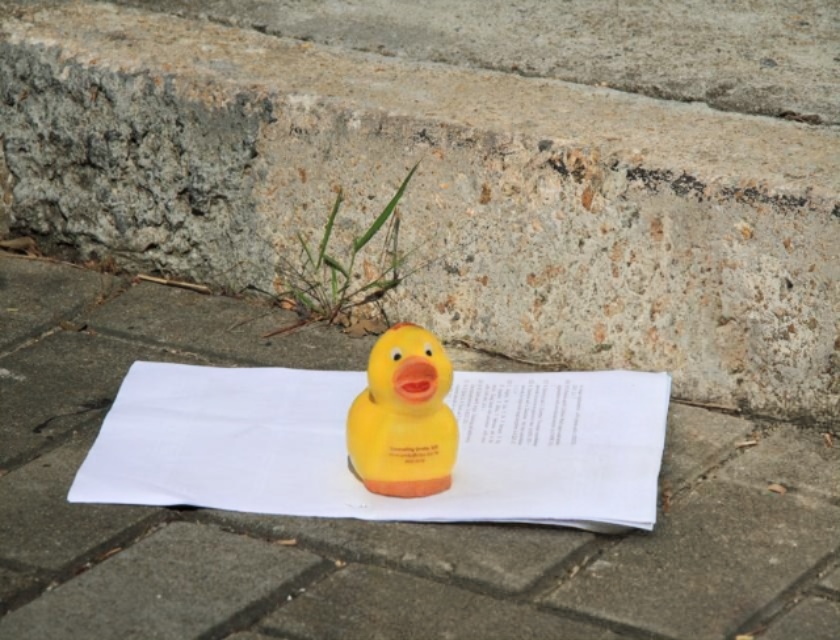}
  \end{minipage}
     \begin{minipage}[t]{0.19\linewidth} 
    \centering 
    \includegraphics[width=0.62in, height=0.62in]{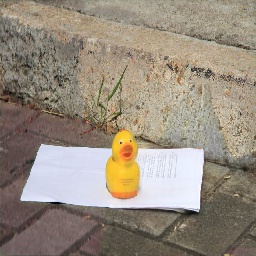}
  \end{minipage} 
   \begin{minipage}[t]{0.19\linewidth} 
    \centering 
    \includegraphics[width=0.62in, height=0.62in]{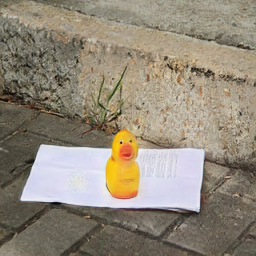}
  \end{minipage} 
    \begin{minipage}[t]{0.19\linewidth} 
    \centering 
        \includegraphics[width=0.62in, height=0.62in]{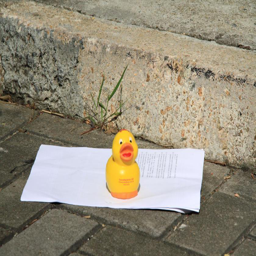}
  \end{minipage} 
  \caption{Results of Task III (Joint shadow/reflection/watermark removal). Row 1-2 show the results of Version one (V1) and Row 3-4 present the results of Version two (V2). We compare BIDeN to DHAN~\cite{cun2020towards} and Auto-Exposure~\cite{fu2021auto} on shadow removal task. Though the performance of BIDeN is constrained by the generality of BIDeN and BID training setting, BIDeN still efficiently removes the shadow and shows competitive results. However, BIDeN suffers from a color shift whereas DHAN and Auto-Exposure better keep the fidelity of the original colors}
  \label{fig:shadow2}
\end{figure}

\begin{figure*}[!htb]
     \centering
     \includegraphics[width = 12cm]
     {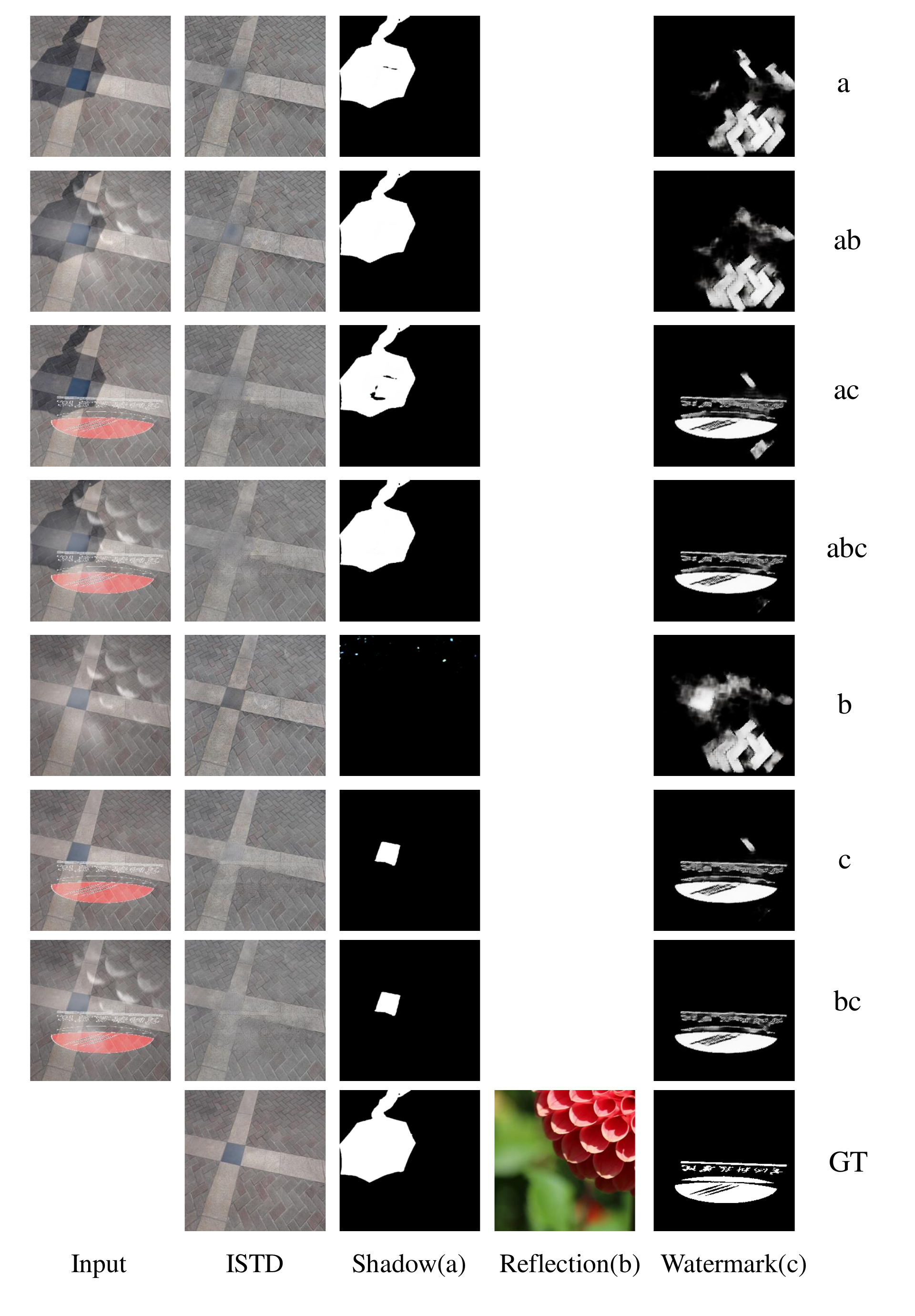}
     \caption{All case results of Task III (Joint shadow/reflection/watermark removal), Version one (V1). ISTD images, shadow masks, and watermark masks generated by BIDeN for all cases. The order of all cases is identical to Table 4 of the main paper. The generated ISTD images suffer color shift, but all shadow/reflection/watermark are efficiently removed for all cases}
     \label{fig:t2s1}
\end{figure*}

\begin{figure*}[!htb]
     \centering
     \includegraphics[width = 12cm]
     {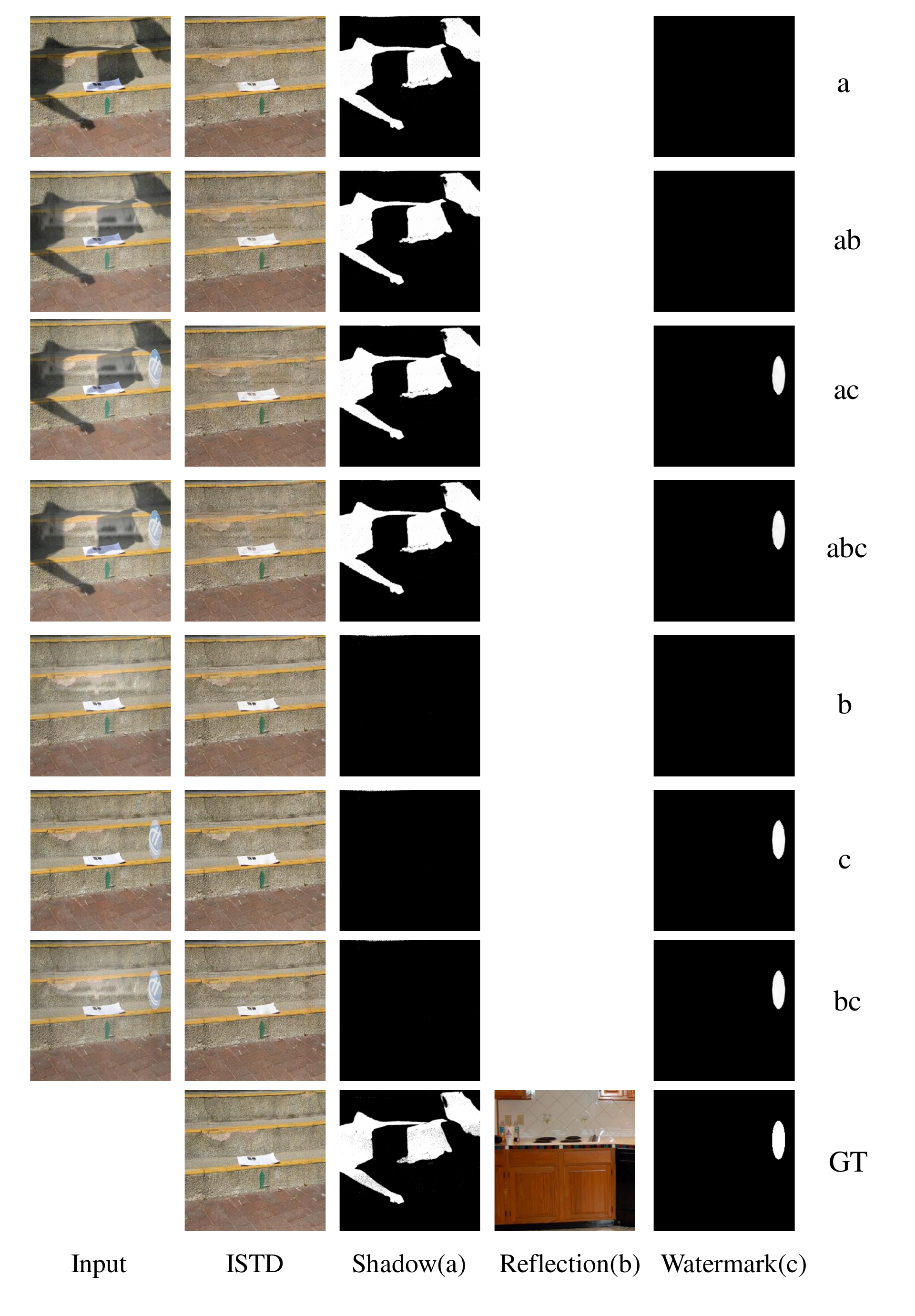}
     \caption{All case results of Task III (Joint shadow/reflection/watermark removal), Version two (V2). ISTD images, shadow masks, and watermark masks generated by BIDeN for all cases. The order of all cases is identical to Table 4 of the main paper. All shadow/reflection/watermark are efficiently removed for all cases}
     \label{fig:t2s2}
\end{figure*}

For Task III (Joint shadow/reflection/watermark removal),
Figure~\ref{fig:shadow2} presents the qualitative comparison between BIDeN and two state-of-the-art baselines~\cite{cun2020towards,fu2021auto}. Though suffering from a color shift, BIDeN still shows visually pleasing, ghost-free shadow removal results.
We also provide additional results of BIDeN for all cases. Results of Version one (V1) and Version two (V2) are shown in Figure~\ref{fig:t2s1} and Figure~\ref{fig:t2s2}.

\clearpage

\end{document}